\documentclass[lettersize,journal]{IEEEtran}
\usepackage{amsmath,amsfonts}
\usepackage{array}
\usepackage[caption=false,font=normalsize,labelfont=sf,textfont=sf]{subfig}
\usepackage{textcomp}
\usepackage{stfloats}
\usepackage{url}
\usepackage{verbatim}
\usepackage{graphicx}
\usepackage{gensymb}
\usepackage{cite}
\usepackage[dvipsnames]{xcolor}
\usepackage{algorithm}
\usepackage{algorithmicx}
\usepackage[noend]{algpseudocode}
\usepackage{makecell}

\usepackage[flushleft]{threeparttable}

\newcommand{\figref}[1]{Fig.~\ref{#1}}   
\newcommand{\tableref}[1]{Table~\ref{#1}}

\newcommand{\rev}[1]{\textcolor{black}{\textnormal{}#1}}
\newcommand{\secondrev}[1]{\textcolor{black}{\textnormal{}#1}}
\newcommand{\thirdrev}[1]{\textcolor{black}{\textnormal{}#1}}

\newcommand{\eqnref}[1]{equation~(\ref{#1})}
\newcommand{\secref}[1]{Section~\ref{#1}}
\newcommand{\algoref}[1]{Algorithm~\ref{#1}}
\newcommand{\reg}[1]{{\textsuperscript{\scriptsize \textregistered} #1}}
\newcommand{\tm}[1]{{\textsuperscript{\tiny TM} #1}}

\begin{document}

\title{CoinFT: A Coin-Sized, Capacitive 6-Axis Force Torque Sensor for Robotic Applications}

\author{Hojung Choi*$^{1}$, Jun En Low*$^{1}$, Tae Myung Huh$^{2}$, Seongheon Hong$^{1}$, Gabriela A. Uribe$^{1}$,\\ Kenneth A. W. Hoffmann$^{1}$, Julia Di$^{1}$, Tony G. Chen$^{3}$, Andrew A. Stanley$^{4}$, and Mark R. Cutkosky$^{1}$

\thanks{ $^1$Stanford University, CA, USA.}
\thanks{ $^2$University of California Santa Cruz, CA, USA.}
\thanks{ $^3$Georgia Institute of Technology, GA, USA.}
\thanks{ $^4$Reality Labs Research, Meta Platforms Inc, WA, USA.}
\thanks{$^{*}$Equal contribution.}
}



\maketitle

\begin{abstract}
We introduce CoinFT, a capacitive 6-axis force\,/\,torque (F\,/\,T) sensor that is compact, light, low-cost, and robust with an average \rev{root-}mean-squared error of 0.16\,N for force and 1.08\,mN$\cdot$m for moment when the input ranges from 0$\sim$14\,N and 0$\sim$5\,N in normal and shear directions, respectively. CoinFT is a stack of two rigid PCBs with comb-shaped electrodes connected by an array of silicone rubber pillars. A microcontroller interrogates the electrodes in different subsets in order to enhance sensitivity for measuring 6-axis F\,/\,T. The combination of features of CoinFT enables various contact-rich robot interactions across different embodiment domains including drones, robot end-effectors, and wearable haptic devices. \rev{We demonstrate the utility of CoinFT through two representative applications: a multi-axial contact-probing experiment in which a CoinFT mounted beneath a hemispherical fingertip measures 6-axis force and torque representative of manipulation scenarios, and a 
force-control task on a drone.} The design, fabrication, and firmware of CoinFT are open-sourced at: {\color{RoyalBlue}\url{https://coin-ft.github.io/}}.
\end{abstract}



\section{Introduction}

\IEEEPARstart{P}{recise} force and torque measurement is vital for robots to perform contact-rich tasks safely and effectively. Tasks such as table wiping \cite{Suh2022}, assembly \cite{Song2021}, or palpating soft tissue \cite{Konstantinova2016} require the application of force and torque within a specific range—sufficient to complete the task but not excessive, as to cause damage or waste energy. Depending on the application and interaction type, robots performing contact-rich tasks come in various forms, including robotic arms \cite{Grice2013}, grippers \cite{Lin2021}, drones \cite{LaVigne2022}, and wearable devices \cite{Yoshida2024}. Therefore, it is highly desirable to develop a compact and low cost solution that can serve a wide variety of applications.

Extensive research has been dedicated to developing 6-axis force\,/\,torque (F\,/\,T) sensors using various transduction methods \cite{Cao2021}. Commercially available sensors also exist, such as the Gamma\texttrademark \,(ATI Industrial Automation), and 6-axis F\,/\,T sensors from MinebeaMitsumi\texttrademark\, or ReSense\texttrademark. However, these sensors are challenging to adopt across different robotic platforms, particularly those requiring both robustness and small, slim form factors. For example, while commercial sensors can be compact and lightweight, they are often too fragile to reliably withstand impacts from drone crashes or incidental contact on an end-effector. Their thickness also adds unnecessary bulk to haptic devices, making them cumbersome. Additionally, the cost of these sensors can be prohibitive, especially when attempting to sensorize multiple fingers of anthropomorphic robot hands such as the Leap hand \cite{shaw2023}. Other sensors that are robust and more affordable tend to be bulky \cite{guggenheim2017}, limiting their applicability in compact platforms. 

In this paper, we introduce CoinFT, a capacitive 6-axis F\,/\,T sensor the size of a US quarter dollar, weighing 2\,g (\figref{fig:SensorDesign} (a)), with a material cost of \rev{less than \$\,11
(\tableref{tab:bom})}. CoinFT combines the aforementioned desirable features,
unlocking various contact-rich robot interactions in unstructured environments.

\rev{We demonstrate CoinFT’s \thirdrev{6-axis} sensing capability through a contact-probing experiment, where a 7-DoF robotic arm applies controlled 6-axis F\,/\,T on a fingertip-mounted CoinFT. This setup demonstrates CoinFT’s suitability for manipulation scenarios while maintaining a compact fingertip-scale form factor.}

Another promising application for CoinFT is on drones performing forceful interactions with the environment. Drones have previously executed contact-rich tasks without force feedback \cite{chen2022} in controlled scenarios or through careful human teleoperation \cite{LaVigne2022}. While force sensors have been demonstrated in such tasks, their size, weight, and fragility make them unsuitable for unstructured environments \cite{guo2023droneTactile}. CoinFT's design addresses these challenges, making it a viable solution for drones operating in environments where large impulses from contact or crashes are common. Its lightweight nature minimally impacts drone operation time, and its compact size facilitates easy integration into even small drones. In this work, we demonstrate the use of CoinFT on a drone for contact modulation through PID force control and the deployment of objects onto environmental surfaces by applying a controlled range of force.


The contributions of this paper are as follows:
\begin{itemize}
    \item The design, fabrication and characterization of CoinFT and an explanation of the principles it utilizes to achieve its performance.
    \item\rev{Experimental validation of CoinFT on multiple robotic platforms, including a contact-probing experiment demonstrating accurate \thirdrev{6-axis} sensing at the fingertip scale and a drone performing contact-force modulation.}
    \item Open-sourcing the design, fabrication, and firmware of CoinFT.
\end{itemize}

\section{Related Work}
\label{sec:RelatedWork}
\subsection{Multi-axial Force Torque Sensors}
Given the role of forces in contact-rich tasks, significant research has focused on developing multi-axial F\,/\,T sensors using diverse transduction methods, including piezoresistive, capacitive, optical, magnetic, and pneumatic techniques \cite{Cao2021}. Among these, piezoresistive approaches have been the most extensively explored, typically involving the attachment of multiple strain gauges to precisely machined structures, commonly designed as cross-beam or parallel structures \cite{Cao2021}. Piezoresistive sensing offers precise measurement capabilities with a high signal-to-noise ratio (SNR), facilitated by strain gauges and circuitry optimized for SNR amplification \cite{Kumar2014}. The availability of compact electronics and advanced machining techniques has also enabled the fabrication of compact sensors, which is why commercially available 6-axis F\,/\,T sensors predominantly utilize piezoresistive technology (e.g. MinebeaMitsumi\texttrademark and ReSense\texttrademark). However, these sensors are often fragile and susceptible to impact, which can result in plastic deformation or even failure of the sensing medium \cite{Stassi2014}. 


\begin{table*}[b]
\centering
\caption{Comparison of representative lower-cost 6-axis force/torque sensors for robotic applications}
\renewcommand{\arraystretch}{1.15}
\setlength{\tabcolsep}{4pt}
\footnotesize
{\color{black} 
\begin{tabular}{|p{2.2cm}|p{1.5cm}|p{1.6cm}|p{0.8cm}|p{2.3cm}|p{2.1cm}|p{2.2cm}|p{1.2cm}|p{1.3cm}|}
\hline
\textbf{Sensor} &
\textbf{Transduction} &
\textbf{Size [mm]} &
\textbf{Mass} &
\textbf{Sensing range} &
\textbf{\makecell[l]{Accuracy \\ ($R^2$ unless \\ noted otherwise)}} &
\textbf{Bandwidth} &
\textbf{Cost} &
\textbf{Robustness} \\ 
\hline
\makecell[l]{
MinebeaMitsumi\texttrademark \\ 6-axis F/T sensor
} &
Piezoresistive &
$\phi$ 9.6 $\times$ $t$ 9 &
3 g &
\makecell[l]{
$F$: 40 N \\ 
\thirdrev{$M$}: 400 mN·m
} &
\makecell[l]{
Full-scale error: \\
$\pm$5\%
} &
1 kHz &
\$1200+ &
No \\ 
\hline
\makecell[l]{
Guggenheim \\ et al.~\cite{guggenheim2017} 
} &
Barometric (MEMS) &
$75\times75\times30$ &
N/A &
\makecell[l]{
$F_{x,y}$: 5 N\\
$F_{z}$: 12 N\\
\thirdrev{$M$}: 3000 mN·m} &
\makecell[l]{
$F_x$: 0.937\\ 
$F_y$: 0.909\\
$F_z$: 0.998\\
$M_x$: 0.984\\
$M_y$: 0.984\\
$M_z$: 0.991} &
50 Hz &
\$20 &
Yes \\ 
\hline
\makecell[l]{
Ouyang \\ et al.~\cite{Ouyang2020}
} &
Vision (Fiducial tracking) &
$36\times31\times51$ &
N/A &
\makecell[l]{
$F_{x,y}$: 2 N\\
$F_{z}$: 1 N\\
\thirdrev{$M_{x,y}$}: 100 mN·m\\
\thirdrev{$M_{z}$}: 4 mN·m} &
\makecell[l]{
$F_x$: 0.991\\
$F_y$: 0.996\\
$F_z$: 0.875\\
$M_x$: 0.997\\
$M_y$: 0.998\\
$M_z$: 0.902} &
25 Hz (sampling) &
\$40 &
Yes \\ 
\hline
\makecell[l]{
Kim \\ et al.~\cite{Kim2020}
} &
Capacitive &
$14\times17\times6.5$ &
2.6 g &
\makecell[l]{$F$: 30 N\\ $M$: 300 mN·m} &
\makecell[l]{
Full-scale error:\\
$Fx$: 0.92\%\\
$Fy$: 0.83\%\\
$Fz$: 1.57\%\\
$Mx$: 2.8\%\\
$My$: 2.1\%\\
$Mz$: 3.4\%} &
\makecell[l]{
200 Hz (sampling)\\
$<$5 Hz \\ (mech. estimated)} &
N/A &
N/A \\ 
\hline
\makecell[l]{
El-Azizi \\ et al.~\cite{elazizi2025}
} &
Optical &
$\phi$ 27 $\times$ $t$ 20 &
16.4 g &
\makecell[l]{$F$: 10 N\\ $M$: 200 mN·m} &
\makecell[l]{
$F_x$: 0.990\\
$F_y$: 0.993\\
$F_z$: 0.560\\
$M_x$: 0.995\\
$M_y$: 0.991\\
$M_z$: 0.923} &
2.5 kHz (sampling) &
\$100 &
Yes \\ 
\hline
\makecell[l]{
Do \\ et al.~\cite{Do2023}
} &
\makecell[l]{Vision \\ (Raw image)} &
32 $\times$ 32 $\times$ 43 &
34 g &
\makecell[l]{
$F_{x,y}$: 5.5 N\\
$F_z$: 11 N\\
\thirdrev{$M_{x,y}$}: 250 mN·m\\
\thirdrev{$M_z$}: 50 mN·m} &
\makecell[l]{
Mean Avg. Error: \\
$F$: 0.41 N\\
\thirdrev{$M$}: 0.387 mN·m\\
} &
30 Hz (sampling) &
\makecell[l]{\$78} &
N/A \\ 
\hline
\makecell[l]{
CoinFT \\ (this work)
} &
Capacitive &
$\phi$ 20 $\times$ $t$ 2 &
2 g &
\makecell[l]{
$F_{x,y}$: 5 N\\
$F_z$: 15 N\\
\thirdrev{$M_{x,y}$}: 100 mN·m\\
\thirdrev{$M_z$}: 80 mN·m} &
\makecell[l]{
$F_x$: 0.983\\
$F_y$: 0.983\\
$F_z$: 0.998\\
$M_x$: 0.988\\
$M_y$: 0.987\\
$M_z$: 0.989} &
\makecell[l]{
360 Hz (sampling)\\
100 Hz (mech.)} &
\$11 &
Yes \\ 
\hline
\end{tabular}
}

\vspace{2mm}
\begin{tablenotes}[flushleft]\footnotesize
\item \thirdrev{*N/A indicates the corresponding specification was not reported in the cited source. Accuracy metrics differ across sources ($R^2$, full-scale error, or mean average error); each entry uses the metric reported in the original work.}
\end{tablenotes}
\label{tab:sensor_comparison}
\end{table*}

Capacitive sensing is another widely adopted approach for designing 6-axis F\,/\,T sensors, valued for its potential to achieve a compact form factor while maintaining high sensitivity \cite{Cao2021}. Compared to piezoresistive sensors, capacitive sensors can be simpler to fabricate. Multi-axis F\,/\,T sensing via capacitive methods is typically achieved in one of two ways. The first uses a single layer of sensing electrodes integrated with specially designed structures that deform to produce distinct signal patterns upon loading \cite{Huh2020, Kim2020, berman2024additively}. \rev{However, such single-layer designs often require complex fabrication processes and specialized equipment for custom machining or microstructuring, which limits scalability.} The second approach employs multi-layer electrode stacks, where each layer responds to specific subsets of forces and torques to generate distinguishable signal patterns \cite{Wu2015}. \rev{While effective, these architectures introduce additional fabrication complexity and can be bulkier, making them less suitable for applications that require compactness. The CoinFT design draws inspiration from multi-layer capacitive architectures but achieves  \thirdrev{6-axis} capability using a single pair of sensing PCBs (printed circuit boards) through a mode-switching and sampling strategy, as discussed in \secref{sec:CoinFT_Design}.}

Optical sensing methods have been employed in the design of multi-axis F\,/\,T sensors by controlling the scattering or reflection of light within the system \cite{Yao2024}. These methods can be broadly categorized into light intensity sensing \cite{almai2018}, sensing based on optical fibers with Fiber Bragg Gratings (FBGs) \cite{Xiong2021}, and vision-based techniques. In traditional optical methods, variations in reflected light intensity at multiple locations are converted into measurements of force and torque \cite{almai2018}. FBG-based sensing employs fibers embedded in areas of higher strain within the sensor body, functioning similarly to strain gauges to precisely measure deformation \cite{Xiong2021}. Vision-based tactile sensing involves using a compact camera to observe the deformation of an elastic medium, often a robot fingertip, upon contact \cite{zhang2022hardware,Do2023, WardCherrier2018TacTip}. A recently popularized approach for force and torque sensing involves monitoring the deformation of specially patterned silicone rubber fingertips and employing data-driven models to infer the applied force and torque \cite{Do2023, Yuan2017, lambeta2024digitizing}. Another method involves tracking the movement of a visual cue with a camera to deduce the load \cite{Li2023vision, Li2021vision}. Alternatively, gel deformation patterns can be remotely observed through an optical fiber, enabling the decoupling of the silicone gel from the sensing camera \cite{baimukashev2020}.

Other, less explored approaches in the design of multi-axial F\,/\,T sensors include piezoelectric \cite{Li2020}, magnetic \cite{black2024, Choi2022}, and pneumatic methods \cite{guggenheim2017}. While piezoelectric effects have been utilized in these sensors, their susceptibility to charge leakage makes them less ideal for measuring static loads. Magnetic sensing techniques, which involve strategically positioning magnets and Hall effect sensors to detect forces and torques across six axes, have also been explored \cite{black2024}. However, magnetic sensing is inherently vulnerable to external noise sources, such as the Earth's magnetic field or nearby ferromagnetic objects. Additionally, similar design principles have been applied in pressure sensors, where differential patterns are created using an array of barometers \cite{guggenheim2017}.

\rev{
Among the various 6-axis F/T sensors reported in prior work, Table~\ref{tab:sensor_comparison} compares representative designs that emphasize compact form factors and relatively low cost, including both research prototypes and commercial devices. CoinFT combines several desirable attributes such as compact and low-profile design, reasonable accuracy across all six axes, mechanical robustness, and affordability, within a single platform. This balance of performance and practicality enables a wide range of robotic applications, including wearable haptic interfaces\cite{Yoshida2024,Sarac2022, winston2025}, dexterous manipulation\cite{chen2025dexforce, choi2026inthewildcompliantmanipulationumift}, and drone-based interaction tasks.
}

\subsection{Contact Based Interaction of Drones}

As described in \secref{sec:DroneApplication}, 
one of the motivating applications of CoinFT in this paper is contact force control with a drone or unmanned aerial vehicle (UAV). Research has addressed the problem of allowing drones to interact with their environment through controlled and sustained contact, such as in perching \cite{Meng2022} or performing tasks that require contact modulation \cite{Aucone2023}. Studies have investigated various applications, such as maintaining a consistent contact force while changing position using hybrid position-force controllers \cite{Bodie2021}, analyzing surface textures by combining load cells with vision-based tactile sensors \cite{guo2023droneTactile}, and attaching objects to surfaces with custom-designed mechanisms \cite{Spieler2023}. However, the load cells used in these studies, typically commercial 6-axis F\,/\,T sensors, present challenges for real-world integration due to their high cost, fragility, and bulky readout electronics. In contrast, CoinFT, with its lightweight, compact design, robustness, and affordability, is ideally suited for drone applications.

\section{\thirdrev{CoinFT Design and Fabrication}}

\subsection{CoinFT Design}
\label{sec:CoinFT_Design}
\begin{figure*}[t]
\centering
	\vspace{1.5mm}
	\includegraphics[width=\textwidth] 
    {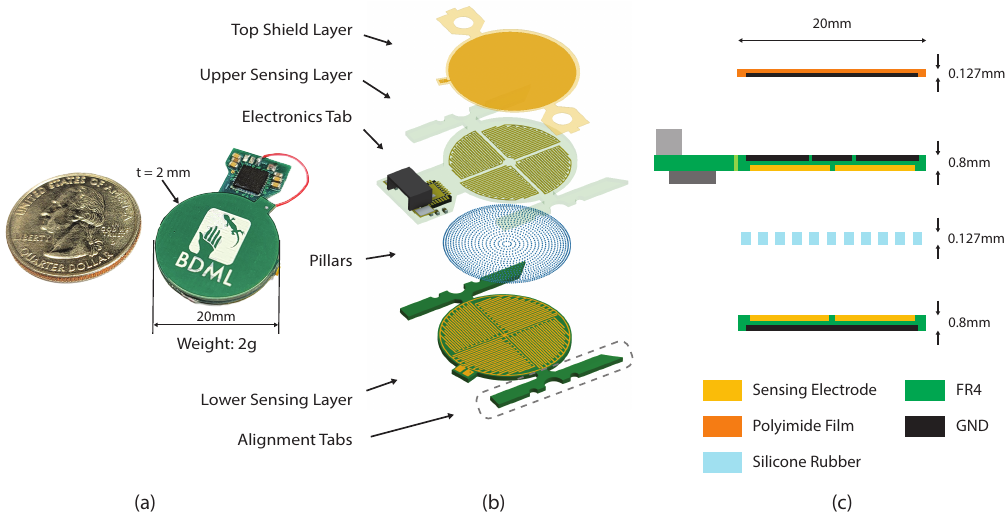}
	\caption{(a) CoinFT is approximately the size of a U.S. quarter-dollar coin. (b) Exploded view showing two rigid PCBs (upper and lower sensing layers\secondrev{, with eight and two individually addressable sensing electrodes, respectively}) connected with an array of silicone rubber pillars. The fPCB top shield layer provides passive shielding. 
    \rev{Alignment tabs are removed after assembly.} (c) \thirdrev{Layer stackup and dimensions} of each layer. The overall thickness is $\approx 2$\,mm.}
	\label{fig:SensorDesign}
\end{figure*}

CoinFT is a capacitive 6-axis force\,/\,torque sensor with a circular sensing area 20\,mm in diameter, an overall thickness of approximately 2\,mm, and a total weight of 2\,g, giving it a footprint comparable to a U.S. quarter-dollar coin (\figref{fig:SensorDesign}\,(a)). As shown in \figref{fig:SensorDesign}\,(b) and (c), it consists of two 0.8\,mm thick rigid PCBs connected by an array of 127\,$\mu$m tall cylindrical pillars cast from silicone rubber (TAP\reg Silicone RTV Mold-Making System). \rev{Solid pillars were selected as they provide favorable sensitivity in both normal and shear directions without compromising bandwidth as some porous dielectrics do\cite{Ham2022, Xia2021}. Silicone rubber was chosen as the pillar material because alternatives such as polyurethane exhibit higher viscoelastic damping and reduced bandwidth \cite{Pichler2023, Cutkosky2014, OSullivan2003, Ltters1997}.} The radial density of pillars increases towards the edge of the PCB for improved strength against delamination. We employ a dense array of sub-millimeter pillars to retain sensitivity. Wider structures will increase effective stiffness of the pillar layer, as discussed in \secref{subsec:Characterization}.


\begin{figure*}[!h]
\centering
	\vspace{1.5mm}
	\includegraphics[width=\textwidth]{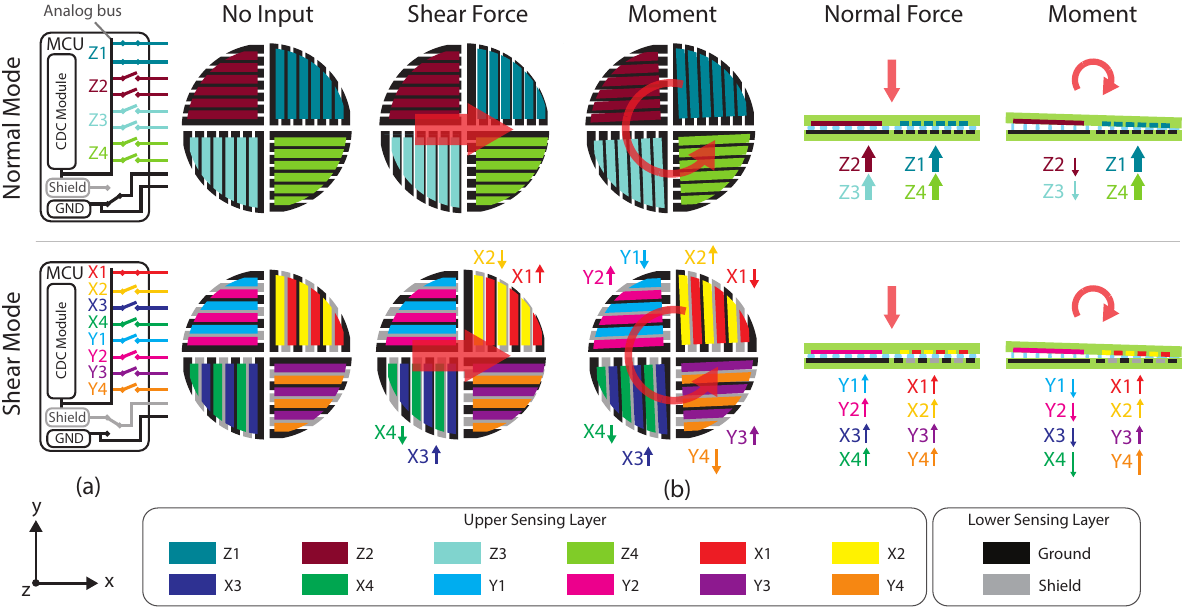}
	\caption{\thirdrev{(a) The microcontroller unit (MCU) firmware switches between two electrode configurations (Normal Mode and Shear Mode) using the analog switches, which connect each electrode to the capacitance-to-digital conversion (CDC) module, to ground (GND), or to an active shield. In the instant depicted in the schematic, \textit{Z1} is being read in Normal Mode while \textit{X1} is being read in Shear Mode. The firmware cycles through all electrodes in both modes to complete one full sampling cycle. 
    (b) The first three columns (no input, shear force, moment) show the resulting capacitance patterns in plan view; the last two (normal force, moment) show them in cross-section. Ground and Shield are the two electrodes of the lower sensing layer (see legend). These configurations produce distinct signal patterns under different force and torque inputs.}}
	\label{fig:SensorWorkingPrinciple}
\end{figure*}

Each PCB is multilayered, with four quadrants of a pair of comb-shaped electrodes \thirdrev{(\figref{fig:SensorDesign}\,(b))} on one side and a single electrode for passive shielding on the other side \thirdrev{(\figref{fig:SensorDesign}\,(c))}. For the upper sensing layer,  the 
\secondrev{quadrants are arranged to create eight 
individually addressable electrodes. The lower layer is also patterned in four quadrants, but with two individually addressable electrodes \thirdrev{(\figref{fig:SensorWorkingPrinciple})}.
} \thirdrev{These electrodes route to the on-board microcontroller (Cypress, PSoC\tm 4100S), whose internal analog switches connect each electrode either to the capacitance-to-digital conversion (CDC) module or to a fixed ground or shield potential (\figref{fig:SensorWorkingPrinciple}\,(a)). By reconfiguring these connections in firmware, a single pair of PCBs becomes sensitive to 6-axis force and torque, as detailed in \secref{subsec:WorkingPrinciple}}. A top shield layer made of a $127 \mu$m thick fPCB (flexible printed circuit board) is added \thirdrev{to provide additional passive shielding from external noise (\figref{fig:SensorDesign}\,(b), (c))}. Each PCB layer has \rev{alignment tabs} on its sides, which are removed during the final stages of fabrication.


An electronics tab (10\,mm $\times$ 11\,mm) attached to the side of the circular sensing area houses the electronics necessary for capacitance sensing \thirdrev{(\figref{fig:SensorDesign}\,(b))}. The microcontroller, with external capacitors, measures and converts capacitance to digital counts using its native capacitance-to-digital conversion (CDC) module with an overall sampling rate of 360\,Hz. Signals can be acquired by other devices through direct serial communication or USB ports through a serial to USB converter, such as a PSoC\tm KitProg2, with a latency of 2\,ms on average.

\subsection{Working Principle}
\label{subsec:WorkingPrinciple}

CoinFT measures 6-axis force and torque by observing the changes in capacitance signal patterns between the upper sensing layer and the lower sensing layer (\figref{fig:SensorDesign}). Capacitance is expressed as:
\begin{equation}
\label{eq:capacitance}
C = \frac{\varepsilon A}{d}
\end{equation}
\noindent where \( C \)  is capacitance, \( \varepsilon \) is the dielectric constant, \( A \) is the overlapping area of the electrodes, and \( d \) is the distance between the electrodes. The microcontroller firmware is programmed to switch the pair of rigid PCBs between two different electrode configurations every sampling sequence: Normal Mode and Shear Mode. \thirdrev{As shown in \figref{fig:SensorWorkingPrinciple}\,(a), each electrode is wired to a set of analog switches inside the microcontroller. Toggling these switches connects an electrode to the CDC module through the analog bus, ties it to ground, or drives it as an active shield. Selecting a different switch pattern therefore reconfigures the same physical electrodes into a different logical sensing arrangement with no change in hardware, a technique similar to that employed by Huh et al. \cite{Huh2020}.}

In Normal Mode, electrodes in each quadrant in the upper sensing layer are internally connected to function as a single electrode, effectively reducing the eight electrodes to four: \textit{Z1}, \textit{Z2}, \textit{Z3}, and \textit{Z4} (\figref{fig:SensorWorkingPrinciple}). \thirdrev{The two electrodes in the lower sensing layer are both reconfigured to connect to \textit{ground}, leaving the \textit{shield} inactive in this mode.} \thirdrev{At the instant depicted in \figref{fig:SensorWorkingPrinciple}\,(a) for Normal Mode, the \textit{Z1} electrodes in the upper right quadrant are tied together and read with respect to ground; the other electrodes are disconnected. Subsequently the \textit{Z2}, \textit{Z3} and \textit{Z4} electrodes will be read in the same way to complete Normal Mode readings for $F_z$, $M_x$ and $M_y$.}

Referring to \eqref{eq:capacitance}, this design allows the Normal Mode to be sensitive to inputs that change \( d \), such as \( F_z \), \( M_x \), and \( M_y \), but not to changes in \( A \), which result from \( F_x \), \( F_y \), and \( M_z \). Under \( F_z \), all Normal Mode electrodes (\textit{Z1}, \textit{Z2}, \textit{Z3}, and \textit{Z4}) show a rise in signal, while under \( M_x \) or \( M_y \), a differential pattern between pairs of Normal Mode electrodes is created (\figref{fig:SensorWorkingPrinciple} (b)).

In Shear Mode, electrodes in the upper sensing layer are kept as independent sensing electrodes (\( X1 \), \( X2 \), \( X3 \), \( X4 \), \( Y1 \), \( Y2 \), \( Y3 \), and \( Y4 \)), while the electrodes in the lower sensing layer are configured such that the \textit{ground} and \textit{shield} electrodes form an alternating pattern (\figref{fig:SensorWorkingPrinciple}). 
The \textit{shield} electrode actively mitigates interference from stray capacitance by being excited to the same electrical potential as the sensing electrodes, an innate capability of PSoC\tm 4100S. This design allows Shear Mode to be sensitive primarily to shear inputs such as \( F_x \), \( F_y \), and \thirdrev{\( M_z \)}. Under a shear load, the relative position of the upper sensing layer and the lower layer shifts laterally, changing \(  A \) between a sensing electrode and \textit{ground} electrode. Due to the negligible capacitance formation between a sensing electrode and a \textit{shield} electrode, the relative position change results in a differential signal pattern in Shear Mode. \thirdrev{At the instant depicted in \figref{fig:SensorWorkingPrinciple}\,(a) for Shear Mode, the \textit{X1} electrodes are read with respect to the alternating ground and shield lines, while the remaining electrodes are disconnected. The \textit{X2}–\textit{X4} and \textit{Y1}–\textit{Y4} electrodes are then read in the same way to complete the Shear Mode scan.}

The comb-shaped electrodes are populated in a way that allows the first and third quadrant of the Shear Mode to be sensitive to loads in the \textit{X} direction, while the second and fourth quadrants are sensitive to the \textit{Y} direction. Under torsion \thirdrev{(\( M_z \))}, all four quadrants in Shear Mode show a differential signal pattern. Unlike Normal Mode configuration where there is negligible cross-modal sensitivity by design, there is cross-coupling in  Shear Mode configuration. \rev{This is resolved in our calibration, as}  discussed in detail in \secref{subsubsec:Calibration}. The unique signal pattern generated from the Normal and Shear Mode configurations upon loads in different axes makes CoinFT a 6-axis F\,/\,T sensor.

\subsection{Sensor Fabrication}

\begin{figure}[!b]
\centering
	\vspace{1.5mm}
	\includegraphics[width=3.1in]{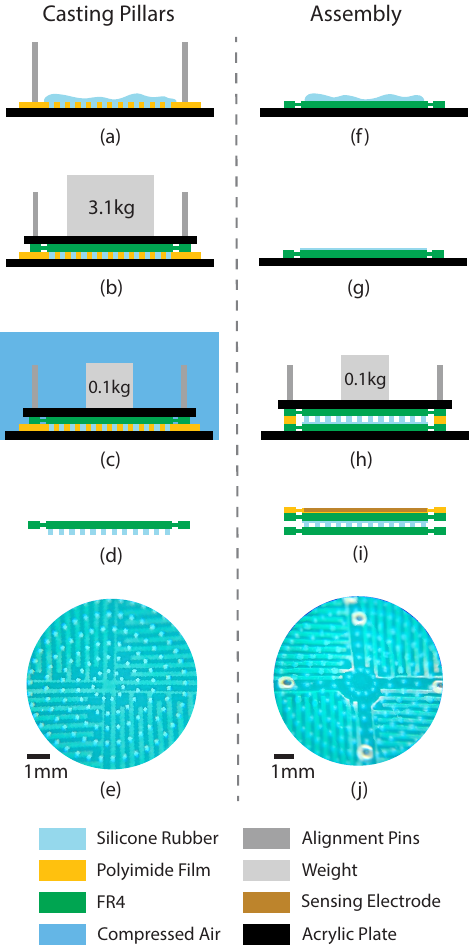}
    \vspace{+1pt}
	\caption{\rev{The fabrication process of CoinFT. (a) Fresh uncured silicone is spread on a UV laser cut mask that is placed on an acrylic plate with alignment pins. (b) A primed lower sensing layer PCB and an acrylic plate for equal pressure distribution is stacked with a 3.1\,kg weight. (c) With a 0.1\,kg weight, the assembly is cured inside a pressurized chamber. (d),(e) Once the mask and acrylic plates are removed, the pillar layer is complete. (f) On a primed upper sensing layer PCB, fresh uncured silicone is spread. (g) The silicone layer is made thin and uniform through spin-coating. (h) The lower sensing layer with pillars is assembled with the upper sensing layer  through precise distance control by adding spacers. A 0.1\,kg weight and an acrylic plate for pressure distribution are added. (i) The top shield layer is attached using an adhesive. (j) A horizontal cross section of the pillar layer shows desirable bonding of the two PCB layers.}} 
	\label{fig:Fabrication}
	\vspace{-0pt}
\end{figure}

CoinFT is fabricated using the process illustrated in \figref{fig:Fabrication}. A 127\,$\mu$m polyimide film pillar mask cut by a UV laser cutter (DPSS Lasers Inc., Samurai UV Marking System) is placed on an acrylic plate using dowel pins for alignment. Vacuum degassed uncured silicone rubber (TAP\reg Silicone RTV Mold-Making System) is smeared on the surface of the mask to fill the cavities for pillars (\figref{fig:Fabrication} (a)). Then, the lower sensing layer PCB, primed using DOWSIL\tm PR-1200 RTV Primer for enhanced adhesion, is aligned and placed on the pillar mask. Another acrylic plate is placed on the assembly for equal pressure distribution and a 3.1\,kg mass is added for 3 minutes to minimize the thickness of the base layer of the silicone pillars (\figref{fig:Fabrication} (b)). The 3\,kg mass is removed, and the pillar assembly, with a 100\,g mass added, is cured inside a pressure chamber for at least 12 hours at approximately 414\,kPa to compress any remaining micro-bubbles inside the silicone rubber (\figref{fig:Fabrication} (c)). Once the acrylic plates and the pillar mask are removed, the pillar layer is complete (\figref{fig:Fabrication} (d), (e)).

\begin{table}[t]
\centering
\caption{Bill of Materials (BOM) for CoinFT}
\renewcommand{\arraystretch}{1.15}
\setlength{\tabcolsep}{5pt}
{\color{black} 
\begin{tabular}{|l|c|c|c|}
\hline
\textbf{Item} & \textbf{Quantity} & \textbf{Unit Cost [\$]} & \textbf{Total Cost [\$]} \\ 
\hline
PSoC 4100S & 1 & 3.25 & 3.25 \\ 
\hline
Upper sensing Layer PCB & 1 & 2.94 & 2.94 \\ 
\hline
Lower sensing layer PCB & 1 & 0.76 & 0.76 \\ 
\hline
Shield PCB & 1 & 2.09 & 2.09 \\ 
\hline
SMD electronics & 6 & 0.20 & 1.20 \\ 
\hline
Molex connector & 1 & 0.69 & 0.69 \\ 
\hline
\textbf{Total} &  &  & \textbf{10.93} \\ 
\hline
\end{tabular}
}
\label{tab:bom}
\end{table}

On the upper sensing layer PCB, after the microcontroller and the electrical components are hand-soldered, the sensing area is primed and applied with vacuum degassed uncured silicone rubber (\figref{fig:Fabrication} (f)). Then, the silicone layer is spin-coated at 5000\,rpm for 75 seconds to achieve a uniform and thin layer (\figref{fig:Fabrication} (g)). The pillar layer is attached to the upper sensing layer PCB with a 203\,$\mu$m polyimide film spacer in between, to control the distance between the two PCBs. A 100\,g mass and an acrylic plate are placed on top of the assembly to apply equally distributed pressure, ensuring that the pillars and the spun silicone layer are in contact (\figref{fig:Fabrication} (h)). The assembly is then cured inside a 65\,\degree C oven for 72 hours. After the alignment pins and spacers are removed, the top shield layer fPCB is adhered to the other side of the upper sensing layer PCB using an adhesive (Henkel Corporation, Loctite 401 Instant Adhesive) (\figref{fig:Fabrication} (i)). A horizontal cross-section of the pillar assembly under a microscope is shown in \figref{fig:Fabrication} (j).
The total cost of all components is \rev{approximately \$11, without scaled production (\tableref{tab:bom}).}

\section{\thirdrev{CoinFT Characterization}}

\thirdrev{We characterize CoinFT in four stages. We first develop a finite element analysis (FEA) model of the pillar array and validate it against measured normal force response (\secref{subsec:Characterization}). We then calibrate the sensor and report its 6-axis accuracy (\secref{subsubsec:Calibration}). We next examine how the pillar geometry can be tuned to trade sensitivity against dynamic range for different applications (\secref{subsec:tuning}). Finally, we characterize operating behavior relevant to deployment, such as mechanical bandwidth, thermal drift and its compensation, and robustness to impact (\secref{subsec:operating_characteristics}).}

\subsection{\thirdrev{Modeling}}

\label{subsec:Characterization}

\begin{figure*}[!tp!]
\centering
	\vspace{1.5mm}
	\includegraphics[width=7in]{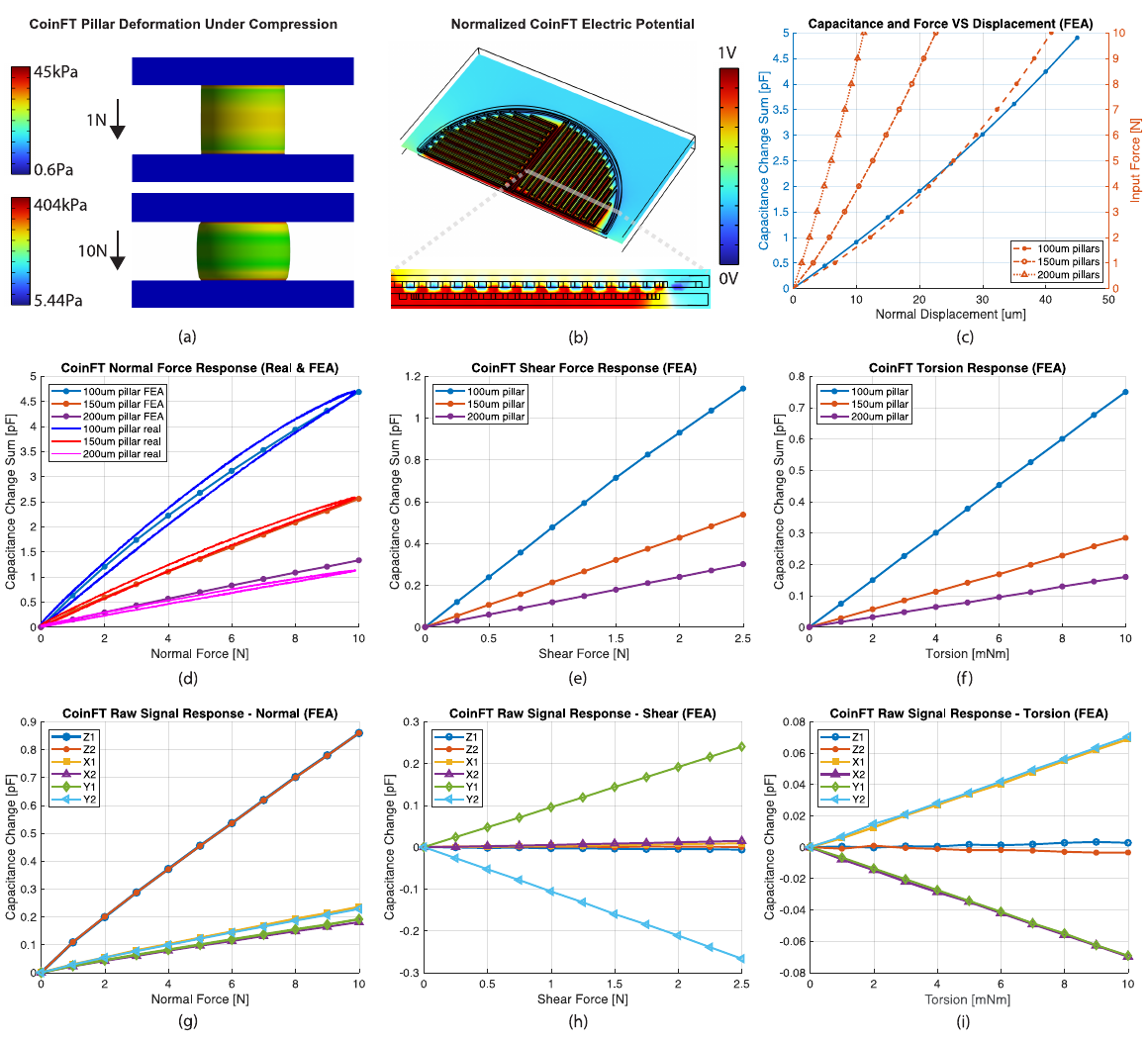}
	\caption{\secondrev{CoinFT characterization using FEA. (a) Structural modeling in FEA. (b) Electrostatics modeling in FEA. (c) Raw capacitance change with displacement and corresponding compressive force. Full CoinFT response with varied pillar diameter under (d) normal force (real samples \& FEA), (e) shear force (FEA), and (f) torsion (FEA). Raw capacitance response of a half of CoinFT (g) under normal force, (h) shear force, and (i) torsion.}}
	\label{fig:SensorCharacterization_FEA}
	\vspace{-3pt}
\end{figure*}

A quantitative analysis using FEA and real sample testing has been conducted to understand the behavior of CoinFT. The FEA involved a structural analysis using ANSYS \secondrev{(\figref{fig:SensorCharacterization_FEA} (a))} to investigate the deformation of the pillar array under different loading conditions, and an electrostatics analysis in COMSOL \secondrev{(\figref{fig:SensorCharacterization_FEA} (b))}, to study the change in capacitance in the sensing electrodes resulting from pillar deformation. \secondrev{The pillar was modeled using a linear elastic constitutive law with the large deformation option enabled to account for geometric nonlinearities. This choice provided a practical balance between computational cost and model fidelity, as a comparison with a hyperelastic (Neo-Hookean) model showed less than 5\% difference in predicted strain within the operating range.} The Gent model \cite{Gent1958} was applied to convert the Shore hardness of the silicone rubber into Young's modulus. With the assumption that the silicone rubber is incompressible, the effective Young's modulus of each pillar was derived using a disk model of an elastomer bonded between two plates, as modeled by Haddow and Ogden \cite{Haddow1988}, shown in \eqref{eq:effectiveYM}:
\begin{equation}
\label{eq:effectiveYM}
E_e = E (1 + \frac{1}{2} \eta^{-2})
\end{equation}
\noindent where \( E_e \) is the effective Young's modulus, \( E \) is the Young's modulus of the material, and \( \eta \) is the ratio between the height and radius of the pillars. The dielectric constant of the silicone rubber was assumed to be 3.0, a common value for silicone rubber \cite{Cho2021}, while the dielectric constant of air was taken as 1.0. 

In order to verify the reliability of our FEA model, we compared the real and simulated response of CoinFT to normal force (\figref{fig:SensorCharacterization_FEA}\,(d)), where the load for the real sample was given up to 10\,N at 0.1\,Hz. The FEA model closely matched the actual sample response, accurately capturing the stiffening effects due to pillar deformation at higher loads, and effectively representing the behavior of CoinFT \secondrev{(\figref{fig:SensorCharacterization_FEA}\,(c),\,(d))}. \secondrev{The same model was used to study the trends in sensor behavior under shear and torsional loading with varying pillar geometry (\figref{fig:SensorCharacterization_FEA}\,(e),\,(f)).}

\subsection{\thirdrev{Calibration and Accuracy}}
\label{subsubsec:Calibration}

\begin{figure*}[bt!]
\centering
	\vspace{1.5mm}
	\includegraphics[width=7in]{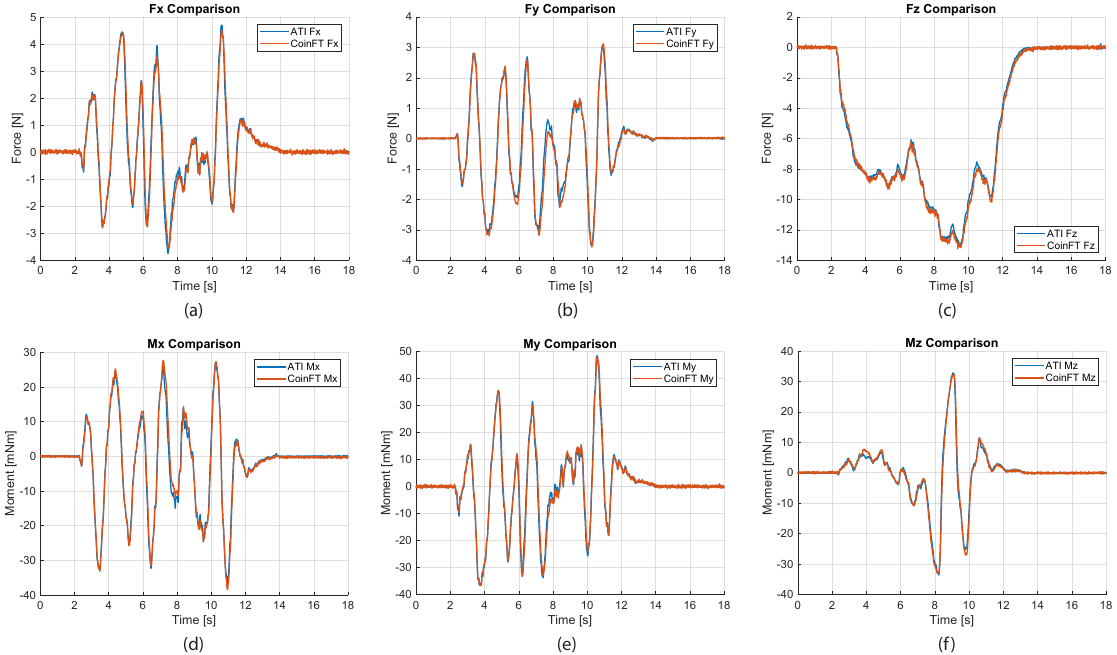}
	\caption{Comparison of sensor readings between CoinFT and Gamma (ATI Industrial Automation) in (a) $F_x$, (b) $F_y$, (c) $F_z$, (d) $M_x$, (e) $M_y$, (f) $M_z$.}
	\label{fig:SensorAccuracy}
	\vspace{-3pt}
\end{figure*}

Based on the working principle from \figref{fig:SensorWorkingPrinciple}, CoinFT produces a distinguishable signal pattern from its 12 sensing electrodes (4 Normal Mode, 8 Shear Mode) under different force and torque loads. The simulated sensor response \secondrev{(\figref{fig:SensorCharacterization_FEA} (g)-(i))} shows the distinguishable signal patterns under ideal alignment and distance of the upper and lower sensing layer PCBs. 

For sensor calibration, CoinFT is \secondrev{rigidly} mounted on a reference sensor (ATI Industrial Automation, Gamma) which provides ground truth force and torque information, while a combination of different loads are given from a human finger. \thirdrev{CoinFT and the reference sensor are loaded simultaneously and measure the same applied wrench, so the data-driven calibration directly maps CoinFT's capacitance signals onto the Gamma's force and torque. Because these loads are applied manually at low rates, far slower than the pillar layer's mechanical response (characterized in \secref{subsec:operating_characteristics}), transient material dynamics have little effect, and the calibration captures a consistent load–signal relationship.}

\rev{Ten trials, each consisting of 35 seconds of data, are used for training, and one additional trial of the same length is reserved for testing. A second-order least-squares regression is employed to map the raw capacitance signals to the corresponding force and torque values. Let \( X \in \mathbb{R}^{24 \times N} \) denote the feature matrix containing the first- and second-order capacitance terms from \( N \) samples, and let \( Y \in \mathbb{R}^{6 \times N} \) denote the corresponding reference forces and torques. The calibration matrix \( A \in \mathbb{R}^{6 \times 24} \) is obtained by solving the normal equation. The calibration results on the test set are shown in \figref{fig:SensorAccuracy}, and the root mean square error (RMSE) and $R^2$ values are provided in \tableref{table:RMSE_CoinFT}. \secondrev{The noise-free resolution of CoinFT, computed as the RMS of the calibrated output under zero-load conditions, is approximately 20\,mN in force and 0.2\,mN$\cdot$m in torque.}}

\begin{table}[tp!]
\centering
\caption{\rev{Accuracy of CoinFT calibration.}}

{\color{black} 
\begin{tabular}{|>{\centering\arraybackslash}m{2.5cm}| >{\centering\arraybackslash}m{0.54cm}| >{\centering\arraybackslash}m{0.54cm}| >{\centering\arraybackslash}m{0.54cm}| >{\centering\arraybackslash}m{0.54cm}| >{\centering\arraybackslash}m{0.54cm}| >{\centering\arraybackslash}m{0.54cm}|}
\hline
\multicolumn{7}{|c|}{\rev{Input Range: 0$\sim$5\,N Normal, 0$\sim$2\,N Shear}} \\
\hline 
    & $F_x$ & $F_y$ & $F_z$ & $M_x$ & $M_y$ & $M_z$ \\ \hline
Normal+Shear\,(RMSE) & 0.038 & 0.040 & 0.050 & 0.317 & 0.306 & 0.231 \\ \hline
Shear\,(RMSE) & 0.039 & 0.040 & 0.115 & 0.343 & 0.323 & 0.244  \\ \hline 
Normal+Shear\,($R^2$) & 0.994 & 0.994 & 0.997 & 0.995 & 0.995 & 0.992 \\ \hline
Shear\,($R^2$) & 0.993 & 0.994 & 0.986 & 0.994 & 0.994 & 0.992  \\ \hline

\hline
\multicolumn{7}{|c|}{Input Range: 0$\sim$14\,N Normal, 0$\sim$5\,N Shear} \\
\hline
  & $F_x$ & $F_y$ & $F_z$ & $M_x$ & $M_y$ & $M_z$ \\ \hline
Normal+Shear\,(RMSE) & 0.186 & 0.134 & 0.186 & 1.114 & 1.322 & 0.810 \\ \hline
Shear\,(RMSE) & 0.174 & 0.129 & 0.266 & 1.169 & 1.587 & 0.867  \\ \hline
Normal+Shear\,($R^2$) & 0.983 & 0.983 & 0.998 & 0.988 & 0.987 & 0.989 \\ \hline
Shear\,($R^2$) & 0.979 & 0.985 & 0.996 & 0.986 & 0.981 & 0.986  \\ \hline

\end{tabular}}

\vspace{2mm}
\begin{tablenotes}[flushleft]\footnotesize
\item \rev{*\secondrev{``Normal + Shear''} includes all 12 electrodes in the calibration process, while the \secondrev{``Shear''} case uses only 8 electrodes from the Shear Mode. Force errors are reported in N, and moment errors are reported in mN$\cdot$m. A CoinFT sample with a pillar diameter of 
$100\,\mu$m was used. RMSE is Root Mean Squared Error.}
\end{tablenotes}
\label{table:RMSE_CoinFT}
\end{table}

The 8 Shear Mode electrodes respond to inputs causing change in distance, \(F_z\), \(M_x\), and \(M_y\), between the electrodes and the \textit{ground} electrode \secondrev{(\figref{fig:SensorCharacterization_FEA}\,(g))}. While it is possible to calibrate the sensor using only the Shear Mode electrodes, the redundancy added by the high signal-to-noise ratio (SNR) of the Normal Mode signals significantly increases calibration accuracy, as shown in \tableref{table:RMSE_CoinFT}. As expected, the increase in accuracy is more significant for \(F_z\), \(M_x\), and \(M_y\).

\subsection{\thirdrev{Sensitivity and Dynamic Range Tuning}}
\label{subsec:tuning}

Different robotic applications require varying sensor dynamic range, sensitivity, and accuracy. For instance, fine control of a fingertip-mounted haptic device \cite{winston2025} requires high sensitivity to small forces within a dynamic range under 4\,N in the normal direction, whereas 
an instrumented probe for ultrasound scans might require sustained loads of 20\,$\sim$\,40\,N \cite{fang2017force} and an instrumented tool for physiotherapy manipulation might require 100\,$\sim$\,200\,N---with correspondingly less emphasis on sensitivity to small forces \cite{chiradejnant2002forces, ren2026multimodalsensingrobotassistedsubtissue}. 
The design of the pillar layer can be adjusted to alter its mechanical properties, meeting the specific requirements of different applications.

The effect of varying pillar diameter on sensor response is illustrated in \secondrev{\figref{fig:SensorCharacterization_FEA}\,(c)-(f)}. Both experimental and FEA results demonstrate a gradual decrease in sensitivity under normal load \secondrev{(\figref{fig:SensorCharacterization_FEA}\,(c),\,(d))}. This reduction in sensitivity is attributed to the stiffening of the pillars as they compress. The constrained lateral deformation of the pillars \secondrev{(\figref{fig:SensorCharacterization_FEA}\,(a))} increases their stiffness, as described in \eqnref{eq:effectiveYM}. Given that the sensitivity of the capacitance change to further compression increases as shown in \eqnref{eq:capacitance} and demonstrated in \secondrev{\figref{fig:SensorCharacterization_FEA}\,(c)}, we note that the stiffening effect is significant. For instance, the quadratic increase of stiffness and sensitivity of capacitance offset each other at lower displacements, resulting in an approximately linear capacitance–force relationship \secondrev{(\figref{fig:SensorCharacterization_FEA}\,(c),\,(d))}. However, the stiffening effect outweighs the increase in capacitance change with higher deformation, explaining the decrease in sensitivity observed in \figref{fig:SensorCharacterization_FEA}\,(d). 

The FEA results also illustrate that the sensitivity to changes in capacitance due to normal and shear forces, as well as torsion, is inversely proportional to the cross-sectional area of the pillar array. Conversely, the dynamic range is proportional to the cross-sectional area of the pillar array.

Additional mechanical parameters that can be adjusted in the pillar layer include the number of pillars, the material selection to alter the Young's modulus, and the arrangement pattern of the pillars. Notably, the torsional stiffness can be fine-tuned while maintaining consistent normal and shear stiffness by modifying the pillar pattern, as this alters the polar moment of inertia without affecting the overall cross-sectional area of the pillar layer. This capability is particularly valuable for applications such as wearable fingertip haptic devices that require high sensitivity of both forces and torque \cite{williams2022}.

\subsection{\thirdrev{Operating Characteristics}}
\label{subsec:operating_characteristics}

\thirdrev{Beyond static accuracy, three characteristics affect CoinFT's behavior in deployment: its mechanical bandwidth, sensitivity to temperature, and tolerance to impact.}

\subsubsection{\thirdrev{Dynamic Response}}
\begin{figure}[t]
\centering
	\vspace{1.5mm}
	\includegraphics[width=3.1in]{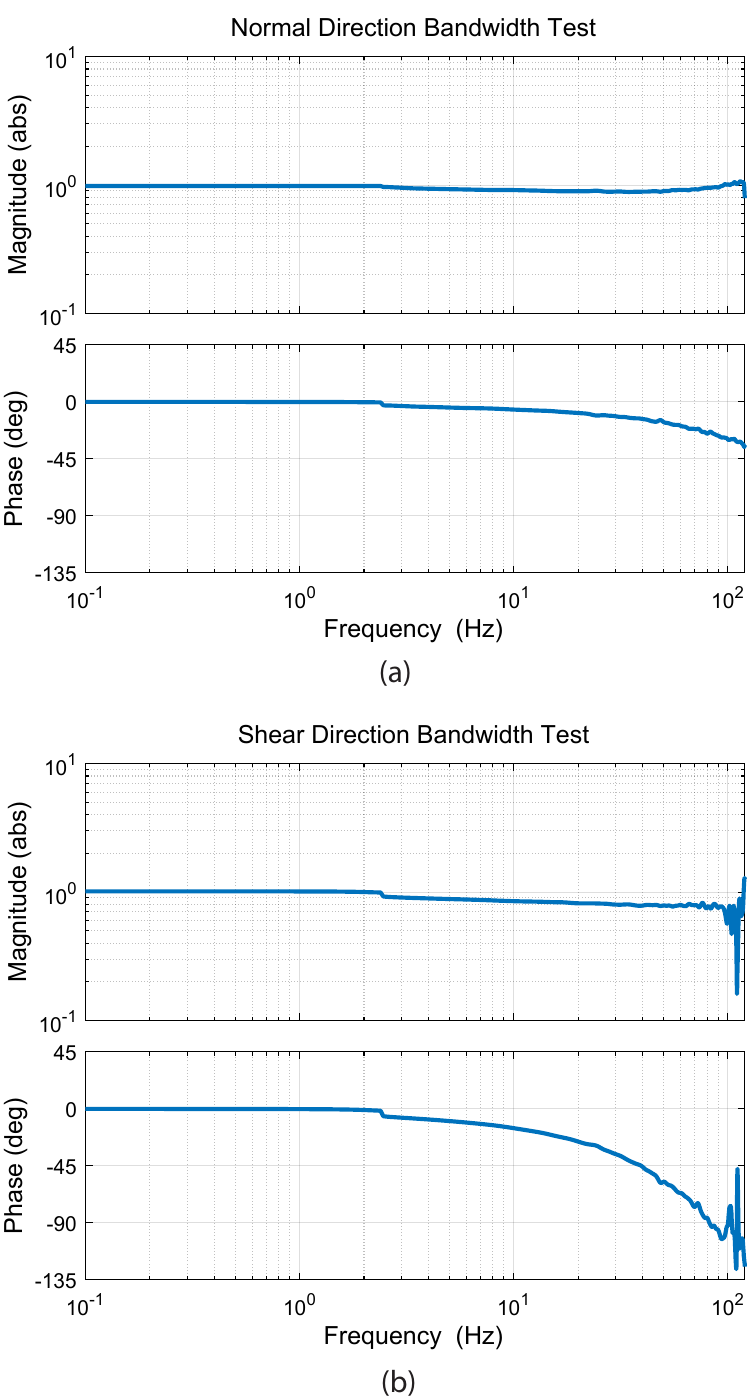}
    \vspace{+1pt}
	\caption{CoinFT frequency response in the (a) normal and (b) shear direction.}
	\label{fig:Bandwidth}
	\vspace{-7pt}
\end{figure}

A frequency response analysis has been conducted to investigate the mechanical bandwidth of CoinFT (\figref{fig:Bandwidth}). A dual-mode muscle lever (Aurora Scientific Inc., 309C) has been used to provide a sinusoidal force sweep either in the normal or shear directions from 0\,Hz $\sim$ 120\,Hz. The force input ranged from 0\,N $\sim$ 2\,N in the normal direction and 0.1\,N $\sim$ 1.1\,N in the shear direction. As shown in \figref{fig:Bandwidth}\,(a), the magnitude response of CoinFT in the normal direction is kept above 70.7\% (-3\,dB) of the input signal across the entire swept frequency. The phase lag is negligible up to 2.5\,Hz, and gradually increases with higher frequency input where, at 120\,Hz, the phase lag is -34.6\,\textdegree. The magnitude response in the shear direction (\figref{fig:Bandwidth}\,(b)) starts to deteriorate at 2.45\,Hz and reaches 70.7\% (-3\,dB) of the input signal at 97\,Hz. At that point, the phase delay is -94.2\,\textdegree. The cause of the phase lag is likely due to the innate hysteresis of silicone rubber material, visible in \figref{fig:SensorCharacterization_FEA}\,(d). These results show that the mechanical bandwidth of CoinFT is approximately 97\,Hz.

\begin{figure}[t!]
\centering
	\vspace{1.5mm}
	\includegraphics[width=3.3in]{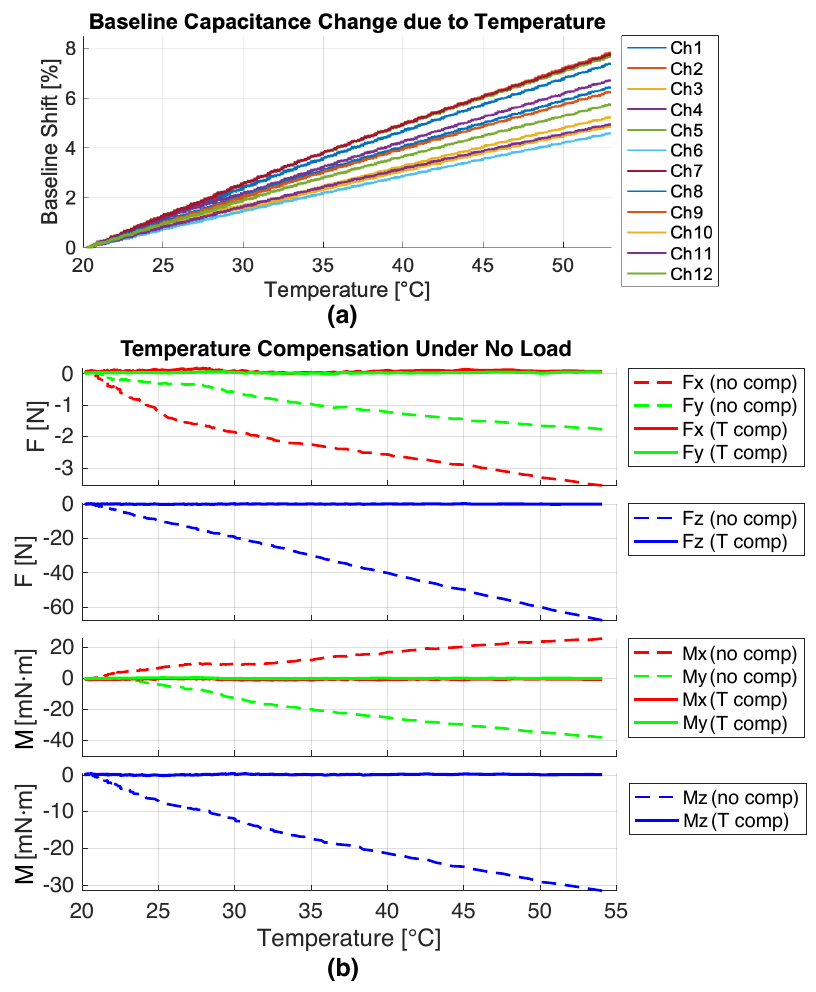}
    \vspace{-7pt}
	\caption{\rev{Effects of temperature variation and its compensation. (a) Drift of raw signals due to temperature change. (b) Temperature compensation using an on-chip temperature sensor.}}
	\label{fig:ThermalEffects}
	\vspace{-7pt}
\end{figure}

\subsubsection{\thirdrev{Thermal Effects}}

\rev{
We investigated the effect of temperature variation and implemented a compensation method using the on-chip temperature sensor of the PSoC. The thermal drift of the raw capacitance signals from all 12 channels was characterized in a temperature-controlled chamber. Across channels, the baseline capacitance drift was approximately 1–3\% per 10\,$^\circ$C. This drift profile was modeled using a second-order polynomial fit, achieving $R^2 > 0.9992$ for all channels. During inference, temperature compensation was performed by subtracting the predicted baseline drift from the raw capacitance signals prior to force/torque estimation. The ablation results under a no-load condition are shown in \figref{fig:ThermalEffects}. With temperature compensation, the residual force error remained below 0.4\,N and torque error below 1.5\,mN·m, whereas without compensation, a clear temperature-dependent drift was observed in the force readings.
}

\subsubsection{\thirdrev{Robustness}}

\begin{figure}[b!]
\centering
	\vspace{1.5mm}
	\includegraphics[width=3.2in]{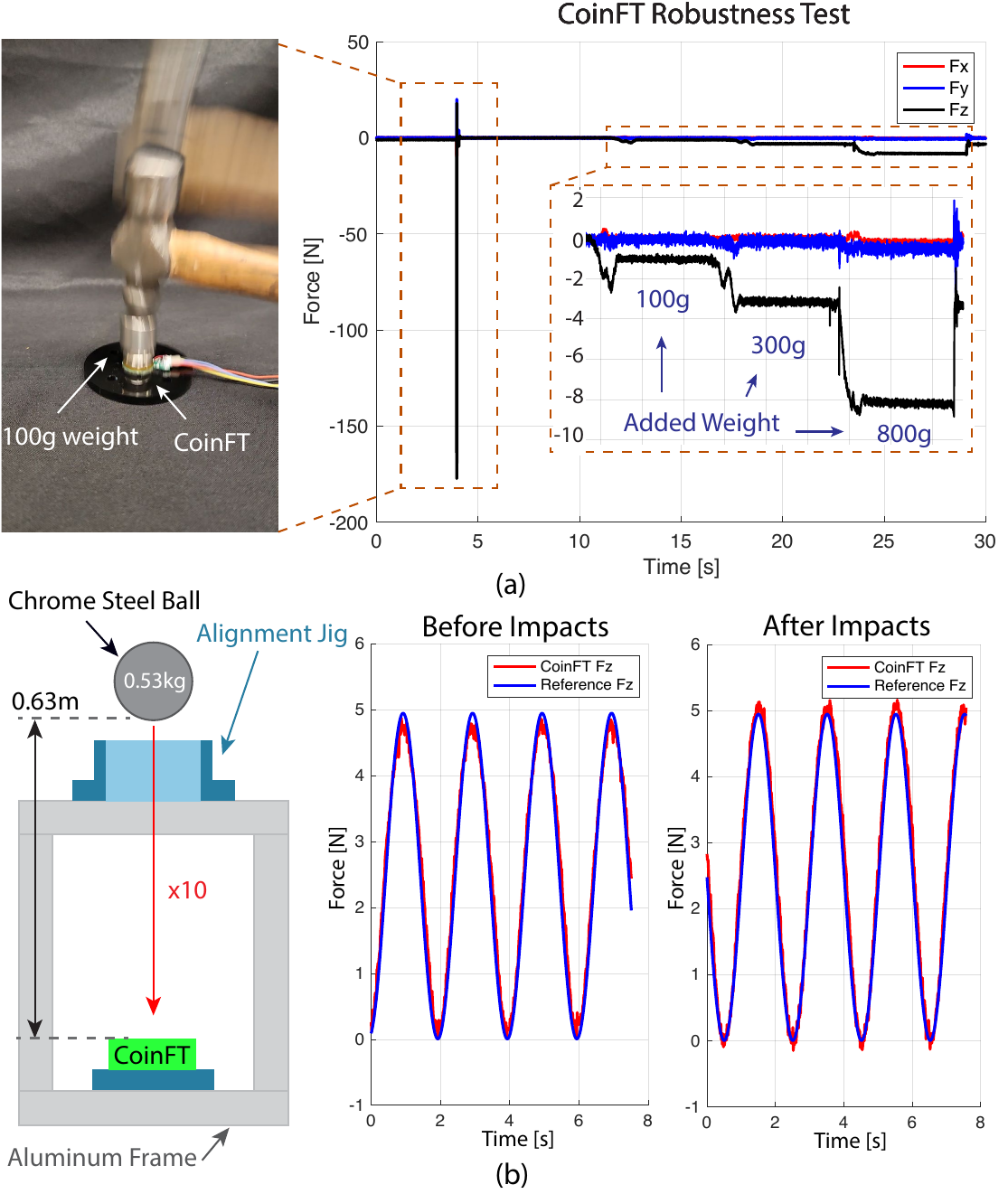}
    \vspace{-7pt}
	\caption{\rev{CoinFT robustness evaluation. (a) CoinFT continues to provide reliable force readings even after an impact from a hammer. (b) The calibration stays consistent after ten impacts from a 0.53\,kg mass dropped at a height of 0.63\,m.}}
	\label{fig:Robustness}
	\vspace{-7pt}
\end{figure}

Many robot applications involve incidental large impulses that can potentially damage commercially available 6-axis F\,/\,T sensors. Due to the compliance of the pillar arrays, CoinFT demonstrates reliable robustness against impacts. As shown in \figref{fig:Robustness}\,(a), CoinFT can accurately measure small loads of 1\,N, 3\,N, and 8\,N even after sustaining an impact from a hammer that causes a force reading of 180\,N. Such an impact would typically damage commercial sensors like the Gamma (ATI Industrial Automation) used for the CoinFT calibration. This level of robustness enables a broader range of robotic interactions beyond controlled environments, making the integration of 6-axis F\,/\,T sensors practical for real-world applications.

\rev{To further quantify the sensor’s impact resistance, a controlled drop test was conducted using a 0.53\,kg chrome steel ball released from a height of 0.63\,m, corresponding to an impact velocity of approximately 3.5\,m\,/\,s and an impact energy of 3.3\,J (\figref{fig:Robustness}\,(b)). Before and after the ten repeated impacts, the CoinFT calibration was validated under a sinusoidal loading profile of 0–5\,N at 0.5\,Hz using a dual-mode muscle lever (Aurora Scientific Inc., 309C). The calibration performance remained consistent, exhibiting a root-mean-squared error of 0.055\,N and an $R^2$ value of 0.982 when compared to the reference force. These results confirm that CoinFT maintains its measurement accuracy and structural integrity under repeated high-energy impacts.}

\rev{While CoinFT exhibits strong robustness under compressive loading, it is less robust in the tensile direction, as excessive tensile stress can cause delamination between the silicone pillars and the PCB substrate. Tensile tests conducted on five CoinFT samples showed an average ultimate tensile strength of 2.84\,MPa (143\,N) with a standard deviation of 1.21\,MPa (61\,N). Increasing the pillar diameter can increase the absolute tensile force the sensor can withstand, while improved bonding strategies, such as surface treatments or adhesive modification, are expected to further enhance the ultimate tensile strength.}

\section{CoinFT Demonstrations}

\rev{While numerous 6-axis force/torque sensors have been developed, as discussed in \secref{sec:RelatedWork}, few achieve a balance of compactness, slim form factor, low cost, accuracy, and robustness. This combination of features enables the broader adoption of F/T sensing across diverse robotic platforms. We demonstrate the versatility of CoinFT through a multi-axial contact probing setup that emulates robot fingertip interactions and an example that motivates its integration on drones.}

\subsection{\rev{Multi-axial Sensing for Robot Manipulators}}

\rev{
Dexterous manipulation tasks, such as object handling or tool use, often require accurate multi-axial force/torque feedback, particularly when interactions require significant contact forces for success\cite{xie2025forcefulroboticfoundationmodels}. For integration on robot fingers, the sensing module must also maintain a compact and slim form factor to avoid drastically altering the hand geometry.}

\begin{figure}[!t]
\centering
	\vspace{1.5mm}
	\includegraphics[width=3.3in]{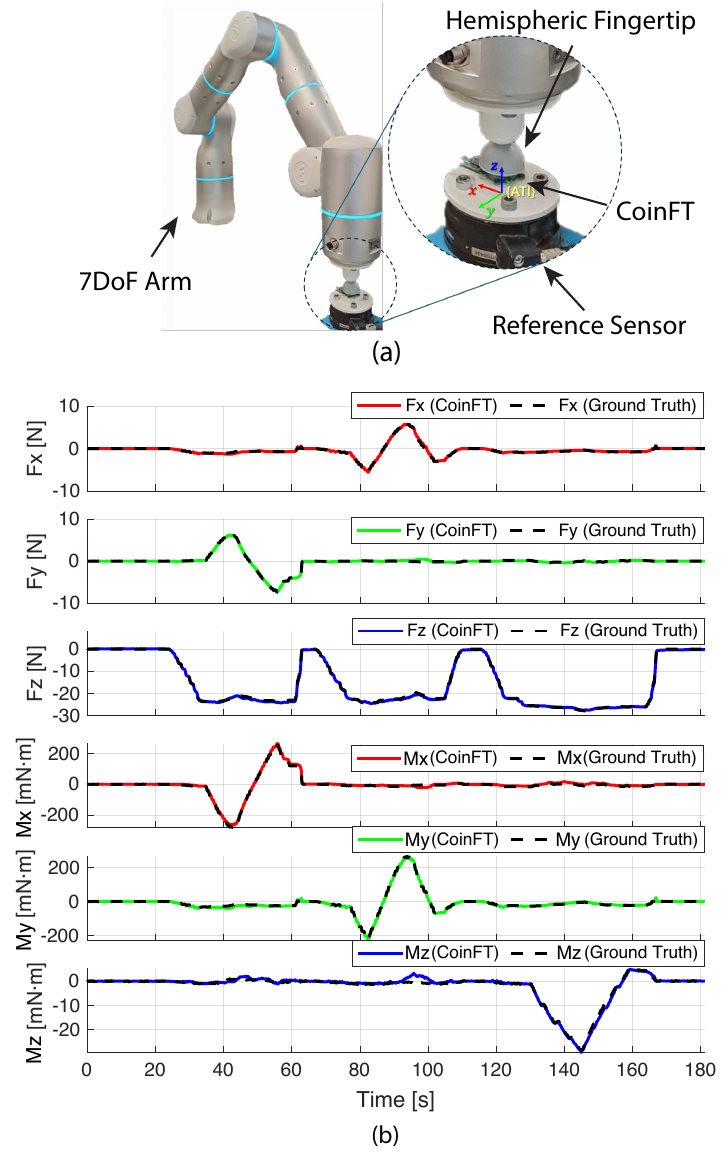}
    \vspace{+1pt}
	\caption{\rev{Multi-axial contact probing experiment. (a) The 7DoF arm applies force and torque in different axes on a hemispherical fingertip equipped with CoinFT, approximating manipulation scenarios. (b) CoinFT and reference sensor reading comparisons. Refer to \tableref{table:Flexiv_CoinFT} for quantitative evaluation.}} 
	\label{fig:Flexiv_CoinFT}
	\vspace{-0pt}
\end{figure}

\rev{
To evaluate CoinFT’s performance in such settings (e.g. as in \cite{chen2025dexforce}), we mounted a CoinFT beneath a robot fingertip with a hemispherical dome and applied varying force and torque along multiple axes using a 7-DoF robotic arm (Flexiv Rizon 4). As shown in \figref{fig:Flexiv_CoinFT}\,(a), a reference force/torque sensor (ATI Gamma) was positioned beneath CoinFT to provide ground-truth measurements. The resulting data (\figref{fig:Flexiv_CoinFT}\,(b)) demonstrate that CoinFT achieves accuracies listed in \tableref{table:Flexiv_CoinFT}.
}

\secondrev{A CoinFT sample with a pillar diameter of 300\,$\mu$m was used in this experiment to extend the operational range to forces common in dexterous manipulation; up to 30\,N in the normal direction and 10\,N in shear. As discussed in \secref{subsec:tuning}, increasing the pillar diameter widens the dynamic range at the cost of reduced sensitivity to smaller forces, while material-dependent characteristics such as mechanical bandwidth remain unchanged. This configuration also increases the tensile limit, improving robustness during manipulation tasks that may involve considerable tensile stresses due to moments.}

\begin{table}[h!]
\centering
\caption{\rev{Multi-axial sensing during contact probing.}}
\renewcommand{\arraystretch}{1.2}
\setlength{\tabcolsep}{6pt}

{\color{black} 
\begin{tabular}{|c|c|c|c|c|c|c|}
\hline
 & $F_x$ & $F_y$ & $F_z$ & $M_x$ & $M_y$ & $M_z$ \\ \hline
RMSE & 0.108 & 0.120 & 0.378 & 5.669 & 3.874 & 0.720 \\ \hline
$R^2$ & 0.994 & 0.995 & 0.999 & 0.993 & 0.996 & 0.986 \\ \hline
\end{tabular}
}

\vspace{2mm}
\begin{tablenotes}[flushleft]\footnotesize
{\color{black} 
\item Input range: 0$\sim$30\,N normal, 0$\sim$7.5\,N shear, 0$\sim$200\,mN·m moment. Forces are reported in N, and moments in mN·m. RMSE denotes the root mean squared error, and $R^2$ represents the coefficient of determination. A CoinFT sample with a pillar diameter of 300\,$\mu$m was used to account for the large input range.
}
\end{tablenotes}

\label{table:Flexiv_CoinFT}
\end{table}

\subsection{Contact Force Modulation of Aerial Vehicles}
\label{sec:DroneApplication}


\rev{Drones offer unique capabilities for performing contact-rich tasks in environments that are difficult or dangerous for humans, such as collecting biological samples from cliffs \cite{LaVigne2022} or cleaning vertical surfaces \cite{Hulaibi2023}. However, most drones lack reliable contact sensing due to the fragility, weight, and high cost of conventional force sensors. Without such feedback, maintaining stable contact or regulating interaction forces is challenging, often resulting in unsafe or inconsistent behaviors.}

\rev{CoinFT addresses these limitations through its compact (20\,mm diameter, 2\,mm thickness), 
lightweight (2\,g, or 6.7\,g including wiring and electronics), and impact-tolerant design. In this section, we demonstrate that a drone equipped with CoinFT and a simple PID-based force controller can stably perform contact interactions, such as pressing and attaching a payload onto a surface. This experiment highlights the ease of integrating CoinFT on aerial platforms.
Given its low cost and mechanical resilience, the sensor can withstand, or be inexpensively replaced after, crashes that are common in drone operations.}

\rev{While only normal force feedback was utilized in this demonstration, CoinFT’s 6-axis sensing capability may enable a broader range of aerial manipulation tasks, such as sliding, scraping, or compliant alignment that would not be feasible without multi-axial sensing.}

\subsubsection{Drone with CoinFT}

\begin{figure}[t]
\centering
	\vspace{1.5mm}
	\includegraphics[width=2.7in]{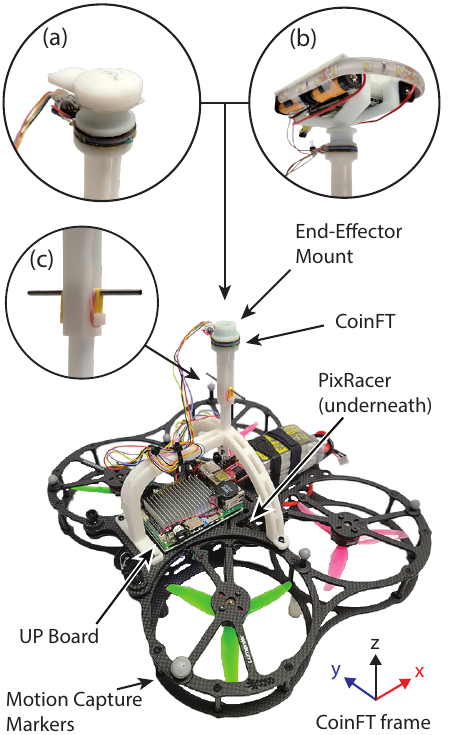}
    \vspace{+1pt}
	\caption{Experimental setup for drones performing contact rich tasks. Different end-effectors can be attached to the CoinFT mount, such as (a) a tip for general contact or (b) a package of electronics to be attached on a surface. Compliance is added in series through a (c) telescoping mechanism with pre-tensioned rubber bands to mitigate impact upon contact.}
	\label{fig:DroneSetup}
	\vspace{-7pt}
\end{figure}

Our drone is a 127\,mm frame quadrotor with an UP Board companion computer and a PixRacer flight controller (\figref{fig:DroneSetup}). CoinFT is attached at the distal tip of a custom mount, which is connected in series with the proximal mount through a telescoping mechanism with rubber bands (\figref{fig:DroneSetup}\,(c)). The compliant element in series helps to reduce impact upon contact and simplify control. CoinFT is connected to the UP Board computer through a USB port. A socket is attached at the top of CoinFT which allows mounting different end-effectors for different tasks, such as a round tip for general contact or a package of electronics (\figref{fig:DroneSetup}\,(a) and (b)). The electronics package consists of an ESP32 microcontroller that uses four 12\,V batteries in parallel to control a strip of LEDs to signal successful package deployment. Pressure sensitive adhesives are attached at the corners on the top side of the package. Since the package is not mechanically attached to the socket, the package will attach to a surface once sufficient pressure provides the necessary adhesion.

\begin{figure}[t]
\centering
	\vspace{1.5mm}
	\includegraphics[width=3.2in]{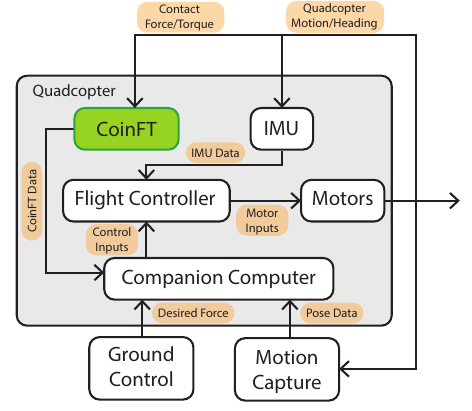}
    \vspace{+1pt}
	\caption{\rev{System diagram of the integrated drone test setup.}}
	\label{fig:SystemDiagram}
	\vspace{-7pt}
\end{figure}

\subsubsection{Attitude Controller for Contact}
To simulate an unstructured environment, we designed a controller for the quadrotor that does not explicitly know the height of a contact surface, thus relying on CoinFT not only to execute force modulation but also to find the contact surface.

Given that the focus here is control with respect to contact, we simplify the navigation aspect of the control problem by providing the drone with accurate estimates of its pose and orientation via an external motion capture system. We then use this as feedback to track a minimum snap trajectory spline \cite{mellinger2011minimum} to a region near the contact surface. We do not provide the drone with the actual height of the contact surface. Instead, our policy will search through a range of heights within which we are confident that the contact surface exists and transform it into a sequence of desired positions ($\boldsymbol{p_{des}}$), velocities ($\boldsymbol{v_{des}}$) and orientations ($\boldsymbol{q_{des}}$), for our controller to track while it searches for contact.

We interface with the PixRacer at the attitude control level, comprising of a commanded normalized thrust force ($\hat{f}_{cmd}$) and a commanded orientation ($\boldsymbol{q_{cmd}}$). We choose this level of control as it allows us to guarantee that the drone is stationary and perpendicular to the contact surface (through commanded orientation) and gives us direct control of the thrust for force modulation.

To account for the drone flight envelope including both in-contact and out-of-contact flight, we use a switching, cascaded PID controller to advance from currently observed states $\begin{bmatrix} \boldsymbol{p_{obs}} & \boldsymbol{v_{obs}} & \boldsymbol{q_{obs}} \end{bmatrix}$ to desired attitude commands $\boldsymbol{u_{cmd}}$. We start by taking the position and velocity error:
\begin{align*}
    \boldsymbol{e_p} = \boldsymbol{p_{obs}} - \boldsymbol{p_{des}} \;, \;\;\;\; \boldsymbol{e_v} = \boldsymbol{v_{obs}} - \boldsymbol{v_{des}}
\end{align*}
from which we compute a desired force vector while accounting for gravity:
\begin{align*}
    \boldsymbol{F_{des}} &= -\boldsymbol{K_p}\boldsymbol{e_p} - \boldsymbol{K_v}\boldsymbol{e_v} - m\boldsymbol{g}
\end{align*}
where \(\boldsymbol{F_{des}}\) is the desired force, \(m\) is the mass of the drone, \(\boldsymbol{g}\) is the gravity vector, and ($\boldsymbol{K_p},\boldsymbol{K_v}$) are positive definite gain matrices that switch between in-contact gains and out-of-contact gains:
\begin{align*}
    \boldsymbol{K_p},\;\boldsymbol{K_v} =
    \begin{cases}
        \boldsymbol{K_p^{oc}},\;\boldsymbol{K_v^{oc}} & \text{if out of contact}\\
        \boldsymbol{K_p^{ic}},\;\boldsymbol{K_v^{ic}} & \text{if in contact}\\
    \end{cases}
\end{align*}

Using the quaternion to rotation matrix to basis vector mapping $\boldsymbol{q_{i}} \Longleftrightarrow \boldsymbol{R_{i}} \Longleftrightarrow \begin{bmatrix} \boldsymbol{x_{i}} & \boldsymbol{y_{i}} & \boldsymbol{z_{i}} \end{bmatrix}$ on both the desired orientation ($\boldsymbol{q_{des}}$) and commanded orientation ($\boldsymbol{q_{cmd}}$), we can compute the latter:
\begin{align*}
    \boldsymbol{z_{cmd}} &= \frac{\boldsymbol{F_{des}}}{||\boldsymbol{F_{des}}||} \\
    \boldsymbol{y_{cmd}} &= \frac{\boldsymbol{z_{cmd}\times}\boldsymbol{x_{des}}}{||\boldsymbol{z_{cmd}\times}\boldsymbol{x_{des}}||} \\
    \boldsymbol{x_{cmd}} &= \boldsymbol{y_{cmd}\times}\boldsymbol{z_{cmd}}
\end{align*}

To get the commanded normalized thrust force we start by getting a desired normalized thrust ($\hat{f}_{des}$). This is done by invoking the quaternion to basis vector mapping to get $\boldsymbol{z_{obs}}$ from $\boldsymbol{q_{obs}}$ and then projecting the desired force vector onto it:
\begin{align*}
    f_{des} &= \boldsymbol{F_{des}} \cdot \boldsymbol{z_{obs}}\\
    \hat{f}_{des} &= k_f {f}_{des}
\end{align*}
where $k_f$ is the thrust normalization coefficient that can be empirically approximated from prior flights. This is then fed into a state machine (\algoref{alg:DroneAlgorithm}) for thrust control that switches modes during contact events and tracks desired contact forces using CoinFT feedback. The two key features of the state machine are a linear search function for the contact force and a PID tracking controller. The former enables transitions when the initial desired contact force is large while the latter not only enables us to hold but also track a changing desired contact force. Combined, these allow for more consistent and safer performance. Given an observed force on CoinFT ($f_{oc}$), desired force on CoinFT ($f_{dc}$) and current time ($t_{cr}$) we perform the 
steps in \algoref{alg:DroneAlgorithm}
to get the commanded normalized thrust. Assuming compressive force, we use the absolute values of the CoinFT reading in the normal direction.

\begin{algorithm}
\caption{Thrust State Machine and PID Controller}
\label{alg:DroneAlgorithm}
\begin{algorithmic}[1]
\State \textbf{Inputs:} \( \hat{f}_{des}, \; f_{oc}, \; f_{dc}, \; t_{cr}\)
\State \textbf{Outputs:} \( \hat{f}_{cmd}, \;  S \)
\State \textbf{Parameters:} Force Increment \( \Delta f \), Proportional Gain \( k_p \), Integral Gain \( k_i \), Derivative Gain \( k_d \), Hold Duration \( T \)
\State \textbf{Initialization:} 
\State \( e_{\text{prev}} \gets 0, \; \int e \, dt \gets 0, \; \hat{f}_{cmd} \gets \hat{f}_{des}, \; S \gets \text{FREE} \)
\If{S is \text{FREE}}
    \State \( \hat{f}_{cmd} \gets \hat{f}_{des} \)
    \If{$f_{oc} > 0$}
        \State \( S \gets \text{SEARCH} \)
    \EndIf
\ElsIf{S is \text{SEARCH}}
    \State \( \hat{f}_{cmd} \gets \hat{f}_{cmd} + \Delta f \)
    \If{$f_{oc} \geq f_{dc}$}
        \State \( f_{hold} \gets \hat{f}_{cmd}, \; t_{prev} \gets t_{cr}, \; t_{0} \gets t_{cr}\)
        \State \( S \gets \text{HOLD} \)
    \EndIf
\ElsIf{S is \text{HOLD}}
    \State \( \Delta t \gets t_{cr}-t_{prev}, \; e_{curr} \gets f_{oc} - f_{dc} \)
    \State \( P \gets k_p e_{curr}, \; I \gets I + k_i e_{curr} \Delta t, \; D \gets k_d \frac{e_{curr} - e_{\text{prev}}}{\Delta t} \)
    \State \( \hat{f}_{cmd} \gets f_{hold} + P + I + D \)
    \State \( e_{\text{prev}} \gets e_{curr} \)
    \If{$t_{cr}-t_{0} \geq T$}
        \State break
    \EndIf
\EndIf
\end{algorithmic}
\end{algorithm}


The CoinFT feedback data lets us use the state machine to dynamically switch the gain matrices $\boldsymbol{K_p},\boldsymbol{K_v}$ where we use the out of contact gains when in FREE state and in contact gains otherwise (\algoref{alg:DroneAlgorithm}). 

\subsubsection{Force Controlled Deployment of Objects on Environment Surfaces}
\begin{figure}[t]
\centering
	\vspace{1.5mm}
	\includegraphics[width=\columnwidth]{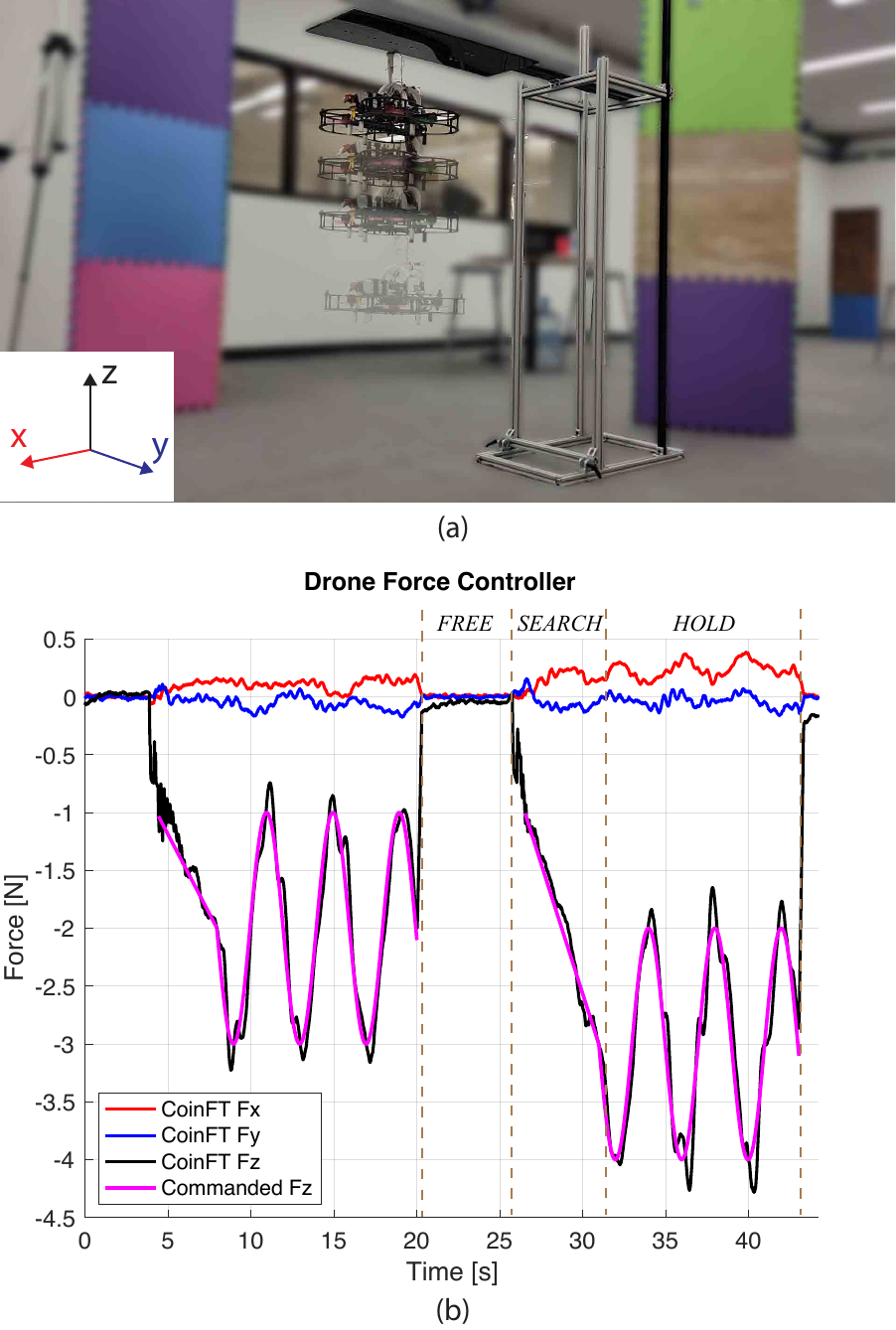}
    \vspace{-17pt}
	\caption{\rev{Results of the force controller test. (a) The drone ascends until contact is made with the test setup, where it initiates force control. (b) CoinFT readings and commanded force.}}
	\label{fig:ForceController}
	\vspace{-7pt}
\end{figure}

To test the designed force controller and perform electronics package delivery tasks, we conducted flight tests. The test setup includes a laser-cut acrylic plate horizontally attached to a structure made of extruded aluminum frames (\figref{fig:ForceController} (a)), simulating an environmental surface. The integrated system consists of a ground control station (PC), motion capture system (OptiTrack, NaturalPoint, Inc.), and a drone equipped with a companion computer, flight controller, IMU, and a CoinFT (\figref{fig:SystemDiagram}). The proposed attitude controller runs on the companion computer \rev{at 20\,Hz}, providing control inputs to the flight controller, which then converts these inputs into PWM signals for the four motors.


The performance of the proposed attitude-based force controller is illustrated in \figref{fig:ForceController}(b). A hemispherical contact tip (\figref{fig:DroneSetup} (a)) is mounted on the drone. \rev{During the \textit{FREE} state, the drone ascends until contact is sensed. In the \textit{SEARCH} state, the thrust is linearly ramped up until a desired contact force is reached, which is initially set by the ground control station. The controller then transitions to the \textit{HOLD} state, where it tracks a pre-programmed sinusoidal force command.} The force controller demonstrates accurate tracking, with an RMS error of 0.1779\,N. \rev{During flight tests, CoinFT showed no noticeable increase in noise or degradation in signal quality despite the potential electromagnetic interference from the motors \cite{watanabe2021measurements}, thanks to its shielding design.}

\begin{figure*}[t]
\centering
	\vspace{1.5mm}
	\includegraphics[width=7in]{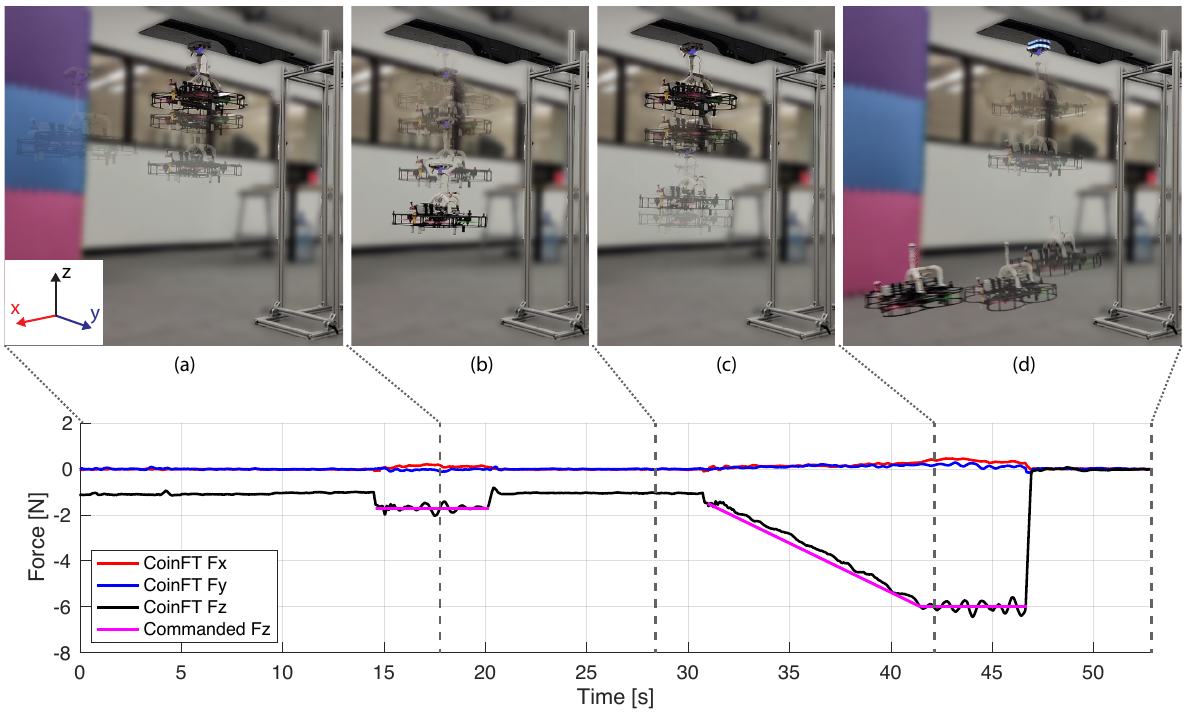}
	\caption{Drone attaching a package of electronics on environment surfaces using force control. (a) The drone gently presses the package on the horizontal surface. (b) After descending, it still feels the weight of the package through CoinFT. (c) The drone presses the package with a larger force on the surface. (d) Upon descent, it no longer feels the weight of the package. It turns on the package and leaves the scene.}
	\label{fig:Demo}
	\vspace{-3pt}
\end{figure*}

Using the force controller, the drone can attach electronics packages to environmental surfaces such as a tree branch by applying a range of force—sufficient to complete the task but not 
sufficient to damage the environment, or expend more energy than required. The electronics package, equipped with pressure-sensitive adhesives and weighing 95\,g (\figref{fig:DroneSetup} (b)), is mounted on the drone. Adequate pressure must be applied for the package to adhere securely to a surface. The drone initially hovers below the horizontal surface and ascends until contact is detected. Upon contact, the contact force ramps up to a commanded value of 0.7\,N, which registers as approximately 1.7\,N in CoinFT due to the package's weight (\figref{fig:Demo} (a)). The drone breaks contact and descends while still sensing approximately 1\,N in CoinFT, detecting that the package has not adhered properly (\figref{fig:Demo} (b)). The drone then autonomously increases the desired contact force to 5\,N (\figref{fig:Demo} (c)). After applying the higher force, the drone no longer senses the package's weight through CoinFT and sends an \textit{ON} command to the electronics package via WiFi, activating the LEDs to signal task success (\figref{fig:Demo} (d)). These controlled interactions are valuable for attaching objects, such as sensor packages, to potentially fragile surfaces like tree branches or for performing general force-controlled tasks, such as cleaning windows on tall buildings.

\section{Discussion \& Future Work}

While this paper \rev{demonstrates CoinFT's capabilities in contact probing or drone platforms,} the sensor's versatility extends to a wide range of other robotic applications. CoinFT has already been used in wearable haptic devices, where it measures and controls the contact force between the device and the human subject. Sarac et al. \cite{Sarac2022} used CoinFT to reveal that people perceive different intensities when the same force is applied in normal versus shear directions. Yoshida et al. \cite{Yoshida2024} employed CoinFT to ensure consistency in user studies by initializing a haptic device tactor based on contact force rather than position. This approach is crucial for accommodating variations in users' body shapes, as position-based initialization does not consistently deliver uniform haptic feedback. CoinFT's slim and compact design also makes it highly suitable for wearable fingertip devices \cite{winston2025}.

CoinFT has also been attached to the fingertips of a commercial anthropomorphic robot hand \cite{chen2025dexforce} for learning force-informed actions from kinesthetic demonstrations. In such a setup, it can sense contact force for controlled, forceful interaction or track the contact point by utilizing force and torque information for intrinsic tactile sensing \cite{Cutkosky2016}. When connected via $I^{2}C$ communication, multiple CoinFTs can be integrated 
to sensorize all fingers at the same 360\,Hz sampling rate,
enhancing the robot hand's ability to perform fine-manipulation tasks. Its low cost and robustness against incidental impacts make CoinFT particularly desirable for robot hands operating in unstructured and cluttered environments.

While the current form factor of CoinFT has been designed to showcase its versatility across various robotic platforms such as drones, robot fingertips, and haptic devices, it can be tailored to meet the specific needs of other applications. For example, CoinFT can be made narrower to sensorize compact medical devices \cite{Chua2023}. The tradeoff in this case would be a decrease in sensitivity due to the reduced sensing area. Conversely, CoinFT could be scaled up to fit the form factor of quadruped or humanoid feet, allowing for the measurement of ground reaction forces (GRF). The larger sensing area in this configuration would enable both higher sensitivity and an expanded dynamic range. This would require a careful selection of design parameters to meet the specifications on sensitivity and dynamic range, as discussed in \secref{subsec:tuning}. \rev{Moreover, the dual-mode electrode-switching strategy used in CoinFT may also inspire future capacitive tactile sensors for enhancing multi-axial sensitivity, particularly those employing programmable capacitance sensing microcontrollers that permit low-level reconfiguration of electrode connections.}

While CoinFT is versatile and suitable for a wide range of robotic applications, it is not intended as a direct replacement for existing commercial sensors like the Gamma (ATI Industrial Automation). The mechanical bandwidth of CoinFT is an order of magnitude lower than that of the Gamma (1\,kHz), making it less capable of detecting highly dynamic contacts. Additionally, the Gamma offers a finer force resolution even with its high dynamic response capabilities.

Other challenges must be considered when using CoinFT. While shielding has been implemented to prevent the coupling of stray capacitance, the surface-mounted electrodes of the chip remain partially exposed, which can result in imperfect shielding. \rev{
While elastomer creep can affect signal, its effect is expected to be small in most applications. The silicone rubber used in CoinFT is a tin-cured RTV elastomer with mechanical properties similar to that in \cite{Cutkosky2014}. As noted then, it has low hysteresis and creep in comparison to other elastomers (e.g. urethanes) of similar stiffness. More detailed characterization of similar silicone materials can be found in \cite{JaneiroArocas2016creep, OSullivan2003, Ltters1997, Wang2011creep}. For example, the material in \cite{Wang2011creep} showed reversible viscoelastic recovery under 
loads up to 0.5 MPa, which would correspond to $\sim$32\,N on CoinFT---suggesting that no permanent creep or damage occurs after repeated loading. Consequently, in applications that do not sustain a large load for long periods of time, mechanical creep is expected to contribute negligibly to signal drift, and recalibration or sensor replacement will not be required.
}

\secondrev{In longer-term deployments, environmental factors beyond temperature, such as humidity, may gradually shift the baseline capacitance signal. In such cases, periodic online re-taring under a known no-load condition can effectively restore measurement accuracy.}

\secondrev{Geometric variations inherent in manual, in-lab fabrication currently necessitate per-sample calibration for each CoinFT prototype. However, transitioning to high-precision mass-manufacturing can improve structural consistency, potentially enabling calibration transfer and reducing the requirement for individual unit characterization.}



The current thin PCB is not entirely rigid and could deform out of plane under significant local pressure, potentially reducing calibration accuracy. This issue, however, can be mitigated by using a slightly thicker PCB or bonding a thin layer of stiff material. Despite the limitations, CoinFT remains valuable for many robotic applications where such high precision and bandwidth in a 6-axis F\,/\,T sensor are not essential, and where robustness and low cost are prioritized.

The proposed attitude control strategy is expected to be effective only in the directions where the drone is fully actuated. Through experimentation, we have observed that precise force control in the lateral direction is challenging with a quadrotor due to the coupling between tilt and lateral displacement. However, CoinFT and the proposed controller can be valuable for modulating lateral contact force when using a fully actuated drone \cite{guo2023droneTactile, Bodie2021}.
\section{Conclusion}
In this paper, we introduced CoinFT, a coin-sized capacitive 6-axis F\,/\,T sensor composed of a pair of specially designed PCBs connected by an array of silicone rubber pillars. Through a comprehensive characterization, CoinFT has demonstrated a unique combination of compactness, robustness, affordability, and reliable performance. These qualities make it an ideal solution for various robotic applications, including drones, robot hands, and wearable haptic devices, particularly in real-world scenarios where large incidental impulses are common.

While CoinFT offers considerable versatility, it is not intended as a replacement for commercially available 6-axis F\,/\,T sensors, which typically offer a larger dynamic range, finer force resolution, and higher bandwidth. Instead, CoinFT excels in applications where such high-end specifications are not a priority, while low cost and scalable deployment are desirable.

We demonstrated the utility of CoinFT by \rev{a multi-axial contact probing experiment emulating fingertip-scale interactions,} and integrating it with a drone to perform force-controlled tasks, such as attaching an electronics package to environmental surfaces. This capability is an example of CoinFT's potential to democratize the use of 6-axis F\,/\,T sensors across a broad range of robotic platforms, enabling more widespread adoption in applications that demand robustness and cost-effectiveness without sacrificing essential functionality. 
To promote this use of CoinFT technology, we have published solid models, fabrication instructions and software as an open-source design at {\color{RoyalBlue}\url{https://coin-ft.github.io/}}.

\section*{ACKNOWLEDGEMENTS}

Meta Reality Labs Research, Toyota Research Institute, and the Kwanjeong Fellowship provided funds to support this work. The authors thank Claire Chen, Kyle Yoshida, Crystal Winston, Mine Sarac, and Alice Wu for exciting and insightful discussions about CoinFT applications.



\vspace{11pt}

\bibliographystyle{IEEEtran}
\bibliography{references}

@article{fang2017force,
  title={Force-assisted ultrasound imaging system through dual force sensing and admittance robot control},
  author={Fang, Ting-Yun and Zhang, Haichong K and Finocchi, Rodolfo and Taylor, Russell H and Boctor, Emad M},
  journal={International journal of computer assisted radiology and surgery},
  volume={12},
  pages={983--991},
  year={2017},
  publisher={Springer}
}

@article{chiradejnant2002forces,
  title={Forces applied during manual therapy to patients with low back pain},
  author={Chiradejnant, Adit and Latimer, Jane and Maher, Christopher G},
  journal={Journal of manipulative and physiological therapeutics},
  volume={25},
  number={6},
  pages={362--369},
  year={2002},
  publisher={Elsevier}
}

@article{berman2024additively,
  title={Additively manufactured micro-lattice dielectrics for multiaxial capacitive sensors},
  author={Berman, Arielle and Hsiao, Kaiwen and Root, Samuel E and Choi, Hojung and Ilyn, Daniel and Xu, Chengyi and Stein, Emily and Cutkosky, Mark and DeSimone, Joseph M and Bao, Zhenan},
  journal={Science Advances},
  volume={10},
  number={40},
  pages={eadq8866},
  year={2024},
  publisher={American Association for the Advancement of Science}
}

@article{lambeta2024digitizing,
  title={Digitizing touch with an artificial multimodal fingertip},
  author={Lambeta, Mike and Wu, Tingfan and Sengul, Ali and Most, Victoria Rose and Black, Nolan and Sawyer, Kevin and Mercado, Romeo and Qi, Haozhi and Sohn, Alexander and Taylor, Byron and others},
  journal={arXiv preprint arXiv:2411.02479},
  year={2024}
}

@inproceedings{elazizi2025,
  title     = {Compact {LED}-Based Displacement Sensing for Robot Fingers},
  author    = {Amr El-Azizi and Sharfin Islam and Pedro Piacenza and
               Kai Jiang and Ioannis Kymissis and Matei Ciocarlie},
  booktitle = {IEEE/RSJ International Conference on Intelligent Robots and Systems},
  year      = {2025}
}

@article{WardCherrier2018TacTip,
  title     = {The TacTip Family: Soft Optical Tactile Sensors with 3D-Printed Biomimetic Morphologies},
  author    = {Benjamin Ward-Cherrier and Nicholas Pestell and Luke Cramphorn and Benjamin Winstone and Maria Elena Giannaccini and Jonathan Rossiter and Nathan F. Lepora},
  journal   = {Soft Robotics},
  volume    = {5},
  number    = {2},
  pages     = {216--227},
  year      = {2018},
  publisher = {Mary Ann Liebert, Inc.}
}

@article{Ham2022,
  title = {Porous Dielectric Elastomer Based Flexible Multiaxial Tactile Sensor for Dexterous Robotic or Prosthetic Hands},
  volume = {8},
  ISSN = {2365-709X},
  number = {3},
  journal = {Advanced Materials Technologies},
  publisher = {Wiley},
  author = {Ham,  Jooyeun and Huh,  Tae Myung and Kim,  Jayoung and Kim,  Jin‐Oh and Park,  Steve and Cutkosky,  Mark R. and Bao,  Zhenan},
  year = {2022},
}

@article{Xia2021,
  title = {Extending Porous Silicone Capacitive Pressure Sensor Applications into Athletic and Physiological Monitoring},
  volume = {21},
  ISSN = {1424-8220},
  number = {4},
  journal = {Sensors},
  author = {Xia,  Yun and Gu,  Hao and Xu,  Lei and Chen,  Xiao Dong and Kirk,  Tim V.},
  year = {2021},
  pages = {1119}
}

@article{Pichler2023,
  title = {Scott Blair Fractional-Type Viscoelastic Behavior of Thermoplastic Polyurethane},
  volume = {15},
  ISSN = {2073-4360},
  number = {18},
  journal = {Polymers},
  author = {Pichler,  Christian and Oberparleiter,  Stefan and Lackner,  Roman},
  year = {2023},
  pages = {3770}
}

@article{Wang2011creep,
  title        = {Study on compressive resistance creep and recovery of flexible pressure sensitive material based on carbon black filled silicone rubber composite},
  author       = {Wang, Luheng and Ma, Fangfang and Shi, Qianshu and Liu, Huanghai and Wang, Xueting},
  journal      = {Sensors and Actuators A: Physical},
  year         = {2011},
  volume       = {164},
  number       = {2},
  pages        = {207--215},
  publisher    = {Elsevier}
}

@article{JaneiroArocas2016creep,
  title        = {Creep analysis of silicone for podiatry applications},
  author       = {Janeiro‐Arocas, Julia and Tarr{\'\i}o‐Saavedra, Javier and L{\'o}pez‐Beceiro, Jorge and Naya, Salvador and L{\'o}pez‐Canosa, Adri{\'a}n and Heredia‐Garc{\'\i}a, Nicol{\'a}s and Artiaga, Ram{\'o}n},
  journal      = {Journal of the Mechanical Behavior of Biomedical Materials},
  year         = {2016},
  volume       = {63},
  pages        = {456--469},
  publisher    = {Elsevier}
}

@article{zhang2022hardware,
  title={Hardware technology of vision-based tactile sensor: A review},
  author={Zhang, Shixin and Chen, Zixi and Gao, Yuan and Wan, Weiwei and Shan, Jianhua and Xue, Hongxiang and Sun, Fuchun and Yang, Yiyong and Fang, Bin},
  journal={IEEE Sensors Journal},
  volume={22},
  number={22},
  pages={21410--21427},
  year={2022},
  publisher={IEEE}
}

@ARTICLE{Huh2020,
  author={Huh, Tae Myung and Choi, Hojung and Willcox, Simone and Moon, Stephanie and Cutkosky, Mark R.},
  journal={IEEE Robotics and Automation Letters}, 
  title={Dynamically Reconfigurable Tactile Sensor for Robotic Manipulation}, 
  year={2020},
  volume={5},
  number={2},
  pages={2562-2569}}

@article{Haddow1988,
  title = {Compression of bonded elastic bodies},
  volume = {36},
  ISSN = {0022-5096},
  number = {5},
  journal = {Journal of the Mechanics and Physics of Solids},
  publisher = {Elsevier BV},
  author = {Haddow,  J.B. and Ogden,  R.W.},
  year = {1988},
  pages = {551–579}
}

@article{Gent1958,
  title={On the Relation between Indentation Hardness and Young's Modulus},
  author={Gent, A. N.},
  journal={Rubber Chemistry and Technology},
  volume={31},
  number={4},
  pages={896--906},
  year={1958}
}

@article{Cho2021,
  title = {Characterization of Mechanical and Dielectric Properties of Silicone Rubber},
  volume = {13},
  ISSN = {2073-4360},
  number = {11},
  journal = {Polymers},
  publisher = {MDPI AG},
  author = {Cho,  Eunyoung and Chiu,  Loraine L. Y. and Lee,  Mitchell and Naila,  Doshina and Sadanand,  Siddharth and Waldman,  Stephen D. and Sussman,  Dafna},
  year = {2021},
  month = jun,
  pages = {1831}
}

@article{Song2021,
  title = {Skill learning for robotic assembly based on visual perspectives and force sensing},
  ISSN = {0921-8890},
  journal = {Robotics and Autonomous Systems},
  publisher = {Elsevier BV},
  author = {Song,  Rui and Li,  Fengming and Quan,  Wei and Yang,  Xuting and Zhao,  Jie},
  year = {2021}
}

@INPROCEEDINGS{Konstantinova2016,
  author={Konstantinova, Jelizaveta and Cotugno, Giuseppe and Dasgupta, Prokar and Althoefer, Kaspar and Nanayakkara, Thrishantha},
  booktitle={IEEE International Conference on Biomedical Robotics and Biomechatronics}, 
  title={Autonomous robotic palpation of soft tissue using the modulation of applied force}, 
  volume={},
  number={},
  year={2016}}

@article{Chua2023,
  title = {A Modular 3-Degrees-of-Freedom Force Sensor for Robot-Assisted Minimally Invasive Surgery Research},
  volume = {23},
  ISSN = {1424-8220},
  number = {11},
  journal = {Sensors},
  publisher = {MDPI AG},
  author = {Chua,  Zonghe and Okamura,  Allison M.},
  year = {2023},
  month = may,
  pages = {5230}
}

@INPROCEEDINGS{Spieler2023,
  author={Spieler, Patrick and Wei, Skylar X. and Li, Monica and Galassi, Andrew and Uckert, Kyle and Kalantari, Arash and Burdick, Joel W.},
  booktitle={2023 IEEE International Conference on Robotics and Automation}, 
  title={PARSEC: An Aerial Platform for Autonomous Deployment of Self-Anchoring Payloads on Natural Vertical Surfaces}, 
  year={2023},
  volume={},
  number={}}

@ARTICLE{Lin2021,
  author={Lin, Michael A. and Thomasson, Rachel and Uribe, Gabriela and Choi, Hojung and Cutkosky, Mark R.},
  journal={IEEE Robotics and Automation Letters}, 
  title={Exploratory Hand: Leveraging Safe Contact to Facilitate Manipulation in Cluttered Spaces}, 
  year={2021}}

@INPROCEEDINGS{Grice2013,
  author={Grice, Phillip M. and Killpack, Marc D. and Jain, Advait and Vaish, Sarvagya and Hawke, Jeffrey and Kemp, Charles C.},
  booktitle={2013 IEEE 13th International Conference on Rehabilitation Robotics}, 
  title={Whole-arm tactile sensing for beneficial and acceptable contact during robotic assistance}, 
  year={2013},
  volume={},
  number={}}

@INPROCEEDINGS{Suh2022,
  author={Suh, H.J. Terry and Kuppuswamy, Naveen and Pang, Tao and Mitiguy, Paul and Alspach, Alex and Tedrake, Russ},
  booktitle={IEEE/RSJ International Conference on Intelligent Robots and Systems}, 
  title={SEED: Series Elastic End Effectors in 6D for Visuotactile Tool Use}, 
  year={2022},
  volume={},
  number={}}

@article{LaVigne2022,
  title = {Collecting critically endangered cliff plants using a drone-based sampling manipulator},
  ISSN = {2045-2322},
  journal = {Scientific Reports},
  publisher = {Springer Science and Business Media LLC},
  author = {La Vigne,  Hughes and Charron,  Guillaume and Rachiele-Tremblay,  Julien and Rancourt,  David and Nyberg,  Ben and Lussier Desbiens,  Alexis},
  year = {2022}
}

@INPROCEEDINGS{Cutkosky2014,
  title = {Dynamic Tactile Sensing},
  ISBN = {9783319030173},
  ISSN = {1610-742X},
  booktitle = {The Human Hand as an Inspiration for Robot Hand Development},
  publisher = {Springer International Publishing},
  author = {Cutkosky,  Mark R. and Ulmen,  John},
  year = {2014}
}

@inproceedings{watanabe2021measurements,
title={Measurements of electromagnetic emission nearby a compact drone},
author={Watanabe, Koh and Aoi, Mai and Komatsu, Misaki and Tanaka, Satoshi and Nagata, Makoto},
booktitle={Asia-Pacific International Symposium on Electromagnetic Compatibility},
year={2021},
organization={IEEE}
}

@article{Ltters1997,
  title = {The mechanical properties of the rubber elastic polymer polydimethylsiloxane for sensor applications},
  volume = {7},
  ISSN = {1361-6439},
  number = {3},
  journal = {Journal of Micromechanics and Microengineering},
  publisher = {IOP Publishing},
  author = {L\"{o}tters,  J C and Olthuis,  W and Veltink,  P H and Bergveld,  P},
  year = {1997},
  month = sep,
  pages = {145–147}
}

@article{OSullivan2003,
  title = {Elastomer rubbers as deflection elements in pressure sensors: investigation of properties using a custom designed programmable elastomer test rig},
  volume = {36},
  ISSN = {1361-6463},
  number = {15},
  journal = {Journal of Physics D: Applied Physics},
  publisher = {IOP Publishing},
  author = {O’Sullivan,  S and Nagle,  R and McEwen,  J A and Casey,  V},
  year = {2003},
  month = jul,
  pages = {1910–1916}
}

@inproceedings{choi2026inthewildcompliantmanipulationumift,
  title     = {In-the-Wild Compliant Manipulation with {UMI-FT}},
  author    = {Hojung Choi and Yifan Hou and Chuer Pan and Seongheon Hong and
               Austin Patel and Xiaomeng Xu and Mark R. Cutkosky and Shuran Song},
  booktitle = {IEEE International Conference on Robotics and Automation},
  year      = {2026}
}

@ARTICLE{winston2025,
  author={Winston, Crystal E. and Choi, Hojung and Jitosho, Rianna and Zhakypov, Zhenishbek and Palmer, Jasmin E. and Cutkosky, Mark R. and Okamura, Allison M.},
  journal={IEEE Transactions on Robotics}, 
  title={Fourigami: A 4-Degree-of-Freedom, Force-Controlled, Origami, Finger Pad Haptic Device}, 
  year={2025},
  volume={41},
  number={},
  pages={4829-4842}}

@inproceedings{ren2026multimodalsensingrobotassistedsubtissue,
  title     = {Multimodal Sensing for Robot-Assisted Sub-Tissue Feature Detection in Physiotherapy Palpation},
  author    = {Tian-Ao Ren and Jorge Garcia and Seongheon Hong and Jared Grinberg and
               Hojung Choi and Julia Di and Hao Li and Dmitry Grinberg and Mark R. Cutkosky},
  booktitle = {Proceedings of the 2026 Design of Medical Devices Conference},
  year      = {2026}
}

@misc{xie2025forcefulroboticfoundationmodels,
      title={Towards Forceful Robotic Foundation Models: a Literature Survey}, 
      author={William Xie and Nikolaus Correll},
      year={2025},
      eprint={2504.11827},
      archivePrefix={arXiv},
      primaryClass={cs.RO},
      note = {arXiv:2504.11827}
}

@ARTICLE{Choi2022,
  author={Choi, Hojung and Brouwer, Dane and Lin, Michael A. and Yoshida, Kyle T. and Rognon, Carine and Stephens-Fripp, Benjamin and Okamura, Allison M. and Cutkosky, Mark R.},
  journal={International Conference on Intelligent Robots and Systems}, 
  title={Deep Learning Classification of Touch Gestures Using Distributed Normal and Shear Force}, 
  year={2022},
  volume={},
  number={},
  pages={3659-3665}}

@ARTICLE{Sarac2022,
  author={Sarac, Mine and Huh, Tae Myung and Choi, Hojung and Cutkosky, Mark R. and Luca, Massimiliano Di and Okamura, Allison M.},
  journal={IEEE Robotics and Automation Letters}, 
  title={Perceived Intensities of Normal and Shear Skin Stimuli Using a Wearable Haptic Bracelet}, 
  year={2022},
  volume={7},
  number={3},
  pages={6099-6106}}

@INPROCEEDINGS{Wu2015,
  author={Wu, X. Alice and Suresh, Srinivasan A. and Jiang, Hao and Ulmen, John V. and Hawkes, Elliot W. and Christensen, David L. and Cutkosky, Mark R.},
  booktitle={IEEE/RSJ International Conference on Intelligent Robots and Systems}, 
  title={Tactile sensing for gecko-inspired adhesion}, 
  year={2015},
  volume={},
  number={}}

@INPROCEEDINGS{Do2023,
  author={Do, Won Kyung and Jurewicz, Bianca and Kennedy, Monroe},
  booktitle={IEEE International Conference on Robotics and Automation}, 
  title={DenseTact 2.0: Optical Tactile Sensor for Shape and Force Reconstruction}, 
  year={2023},
  volume={},
  number={}}

@article{Yao2024,
  title = {Recent progress of optical tactile sensors: A review},
  volume = {176},
  ISSN = {0030-3992},
  journal = {Optics \& Laser Technology},
  publisher = {Elsevier BV},
  author = {Yao,  Ni and Wang,  Shipeng},
  year = {2024},
  pages = {111040}
}

@article{Yuan2017,
  title = {GelSight: High-Resolution Robot Tactile Sensors for Estimating Geometry and Force},
  volume = {17},
  ISSN = {1424-8220},
  number = {12},
  journal = {Sensors},
  publisher = {MDPI AG},
  author = {Yuan,  Wenzhen and Dong,  Siyuan and Adelson,  Edward},
  year = {2017},
  pages = {2762}
}

@ARTICLE{Kim2020,
  author={Kim, Uikyum and Jeong, Heeyeon and Do, Hyunmin and Park, Jongwoo and Park, Chanhun},
  journal={IEEE Robotics and Automation Letters}, 
  title={Six-Axis Force/Torque Fingertip Sensor for an Anthropomorphic Robot Hand}, 
  year={2020},
  volume={5},
  number={4},
  pages={5566-5572}}

@ARTICLE{baimukashev2020,
  author={Baimukashev, Daulet and Kappassov, Zhanat and Varol, Huseyin Atakan},
  journal={IEEE Robotics and Automation Letters}, 
  title={Shear, Torsion and Pressure Tactile Sensor via Plastic Optofiber Guided Imaging}, 
  year={2020},
  volume={5},
  number={2},
  pages={2618-2625}}

@article{chen2025dexforce,
  title   = {{DexForce}: Extracting Force-Informed Actions From Kinesthetic Demonstrations for Dexterous Manipulation},
  author  = {Claire Chen and Zhongchun Yu and Hojung Choi and
             Mark Cutkosky and Jeannette Bohg},
  journal = {IEEE Robotics and Automation Letters},
  year    = {2025},
  volume  = {10},
  number  = {6},
  pages   = {6416--6423},
  doi     = {10.1109/LRA.2025.3568318}
}

@article{Li2020,
  title = {Fault-tolerant measurement mechanism research on pre-tightened four-point supported piezoelectric six-dimensional force/torque sensor},
  volume = {135},
  ISSN = {0888-3270},
  journal = {Mechanical Systems and Signal Processing},
  publisher = {Elsevier BV},
  author = {Li,  Ying-jun and Xu,  Xing-sheng and Wang,  Gui-cong and Cao,  Shu-kun and Chen,  Nai-jian and Sun,  Xuan},
  year = {2020},
  pages = {106420}
}

@ARTICLE{black2024,
  author={Black, David G. and Hossein Hadi Hosseinabadi, Amir and Rangga Pradnyawira, Nicholas and Nogami, Mika and Salcudean, Septimu E.},
  journal={IEEE Transactions on Medical Robotics and Bionics}, 
  title={Low-Profile 6-Axis Differential Magnetic Force/Torque Sensing}, 
  year={2024},
  volume={6},
  number={3},
  pages={992-1003}}

@ARTICLE{Li2021vision,
  author={Li, Wanlin and Alomainy, Akram and Vitanov, Ivan and Noh, Yohan and Qi, Peng and Althoefer, Kaspar},
  journal={IEEE Sensors Journal}, 
  title={F-TOUCH Sensor: Concurrent Geometry Perception and Multi-Axis Force Measurement}, 
  year={2021},
  volume={21},
  number={4},
  pages={4300-4309}}

@ARTICLE{Li2023vision,
  author={Li, Wanlin and Wang, Meng and Li, Jiarui and Su, Yao and Jha, Devesh K. and Qian, Xinyuan and Althoefer, Kaspar and Liu, Hangxin},
  journal={IEEE Robotics and Automation Letters}, 
  title={L$^{3}$ F-TOUCH: A Wireless GelSight With Decoupled Tactile and Three-Axis Force Sensing}, 
  year={2023},
  volume={8},
  number={8},
  pages={5148-5155}}

@ARTICLE{Xiong2021,
  author={Xiong, Li and Guo, Yongxing and Jiang, Guozhang and Zhou, Xinglin and Jiang, Lin and Liu, Honghai},
  journal={IEEE Transactions on Industrial Electronics}, 
  title={Six-Dimensional Force/Torque Sensor Based on Fiber Bragg Gratings With Low Coupling}, 
  year={2021},
  volume={68},
  number={5},
  pages={4079-4089}}

@ARTICLE{almai2018,
  author={Al-Mai, Osama and Ahmadi, Mojtaba and Albert, Jacques},
  journal={IEEE Sensors Journal}, 
  title={Design, Development and Calibration of a Lightweight, Compliant Six-Axis Optical Force/Torque Sensor}, 
  year={2018},
  volume={18},
  number={17},
  pages={7005-7014}}

@ARTICLE{williams2022,
  author={Williams, Sophia R. and Suchoski, Jacob M. and Chua, Zonghe and Okamura, Allison M.},
  journal={IEEE Robotics and Automation Letters}, 
  title={A 4-Degree-of-Freedom Parallel Origami Haptic Device for Normal, Shear, and Torsion Feedback}, 
  year={2022},
  volume={7},
  number={2},
  pages={3310-3317}}

@ARTICLE{Yoshida2024,
  author={Yoshida, Kyle T. and Zook, Zane A. and Choi, Hojung and Luo, Ming and O'Malley, Marcia K. and Okamura, Allison M.},
  journal={IEEE Transactions on Haptics}, 
  title={Design and Evaluation of a 3-DoF Haptic Device for Directional Shear Cues on the Forearm}, 
  year={2024},
  volume={},
  number={}}

@article{Kumar2014,
  title = {Design principles and considerations for the ‘ideal’ silicon piezoresistive pressure sensor: a focused review},
  volume = {20},
  ISSN = {1432-1858},
  number = {7},
  journal = {Microsystem Technologies},
  publisher = {Springer Science and Business Media LLC},
  author = {Kumar,  S. Santosh and Pant,  B. D.},
  year = {2014},
  pages = {1213–1247}
}

@article{Stassi2014,
  title = {Flexible Tactile Sensing Based on Piezoresistive Composites:  A Review},
  volume = {14},
  ISSN = {1424-8220},
  number = {3},
  journal = {Sensors},
  publisher = {MDPI AG},
  author = {Stassi,  Stefano and Cauda,  Valentina and Canavese,  Giancarlo and Pirri,  Candido},
  year = {2014},
  pages = {5296–5332}
}

@ARTICLE{Cao2021,
  author={Cao, Max Yiye and Laws, Stephen and Baena, Ferdinando Rodriguez y},
  journal={IEEE Sensors Journal}, 
  title={Six-Axis Force/Torque Sensors for Robotics Applications: A Review}, 
  year={2021}}

@ARTICLE{chen2022,
  author={Chen, Tony G. and Hoffmann, Kenneth A. W. and Low, Jun En and Nagami, Keiko and Lentink, David and Cutkosky, Mark R.},
  journal={IEEE Robotics and Automation Letters}, 
  title={Aerial Grasping and the Velocity Sufficiency Region}, 
  year={2022}}

@article{Aucone2023,
  title = {Drone-assisted collection of environmental DNA from tree branches for biodiversity monitoring},
  volume = {8},
  ISSN = {2470-9476},
  number = {74},
  journal = {Science Robotics},
  publisher = {American Association for the Advancement of Science (AAAS)},
  author = {Aucone,  Emanuele and Kirchgeorg,  Steffen and Valentini,  Alice and Pellissier,  Loïc and Deiner,  Kristy and Mintchev,  Stefano},
  year = {2023}
}

@article{Meng2022,
  title = {On Aerial Robots with Grasping and Perching Capabilities: A Comprehensive Review},
  volume = {8},
  ISSN = {2296-9144},
  journal = {Frontiers in Robotics and AI},
  publisher = {Frontiers Media SA},
  author = {Meng,  Jiawei and Buzzatto,  Joao and Liu,  Yuanchang and Liarokapis,  Minas},
  year = {2022}
}

@ARTICLE{guggenheim2017,
  author={Guggenheim, Jacob W. and Jentoft, Leif P. and Tenzer, Yaroslav and Howe, Robert D.},
  journal={IEEE/ASME Transactions on Mechatronics}, 
  title={Robust and Inexpensive Six-Axis Force–Torque Sensors Using MEMS Barometers}, 
  year={2017}}

@INPROCEEDINGS{Ouyang2020,
  author={Ouyang, Rui and Howe, Robert},
  booktitle={International Conference on Robotics and Automation}, 
  title={Low-Cost Fiducial-based 6-Axis Force-Torque Sensor}, 
  year={2020},
  volume={},
  number={}}

@incollection{Cutkosky2016,
  comment={inbook does not work},
  title = {Force and Tactile Sensing},
  ISBN = {9783319325521},
  ISSN = {2522-8706},
  booktitle = {Springer Handbook of Robotics},
  publisher = {Springer International Publishing},
  author = {Cutkosky,  Mark R. and Provancher,  William},
  year = {2016},
}

@inproceedings{shaw2023,
  title     = {{LEAP} Hand: Low-Cost, Efficient, and Anthropomorphic Hand for Robot Learning},
  author    = {Kenneth Shaw and Ananye Agarwal and Deepak Pathak},
  booktitle = {Proceedings of Robotics: Science and Systems},
  year      = {2023}
}

@INPROCEEDINGS{Hulaibi2023,
  author={Hulaibi, Amal Al and Wahid, Hawraa and Saif, Salem Al and Sadeq, Jumanah and Esmaeili, Seyed and Kandil, Marwa},
  booktitle={International Conference on Electrical, Communication and Computer Engineering}, 
  title={SkyBot - A Novel Autonomous Window Cleaning Drone}, 
  year={2023},
  volume={},
  number={},
  doi={10.1109/ICECCE61019.2023.10442126}}

@inproceedings{mellinger2011minimum,
  title={Minimum snap trajectory generation and control for quadrotors},
  author={Mellinger, Daniel and Kumar, Vijay},
  booktitle={2011 IEEE international conference on robotics and automation},
  year={2011},
  organization={IEEE}
}

@inproceedings{guo2023droneTactile,
  title     = {Aerial Interaction with Tactile Sensing},
  author    = {Xiaofeng Guo and Guanqi He and Mohammadreza Mousaei and
               Junyi Geng and Guanya Shi and Sebastian Scherer},
  booktitle = {IEEE International Conference on Robotics and Automation},
  year      = {2024}
}

@ARTICLE{Bodie2021,
  author={Bodie, Karen and Brunner, Maximilian and Pantic, Michael and Walser, Stefan and Pfändler, Patrick and Angst, Ueli and Siegwart, Roland and Nieto, Juan},
  journal={IEEE Transactions on Robotics}, 
  title={Active Interaction Force Control for Contact-Based Inspection With a Fully Actuated Aerial Vehicle}, 
  year={2021},
  volume={37},
  number={3},
  pages={709-722},
  doi={10.1109/TRO.2020.3036623}}

\vfill

\end{document}